\documentclass[10pt,twocolumn,letterpaper]{article}

\usepackage{iccv}
\usepackage{times}
\usepackage{epsfig}
\usepackage{graphicx}
\usepackage{amsmath}
\usepackage{amssymb}

\usepackage[utf8]{inputenc} 
\usepackage[T1]{fontenc}    
\usepackage{url}            
\usepackage{booktabs}       
\usepackage{amsfonts}       
\usepackage{nicefrac}       
\usepackage{microtype}      
\usepackage{graphicx}
\usepackage{color}
\usepackage{multirow}
\usepackage{wrapfig}
\usepackage[dvipsnames]{xcolor}
\usepackage{isomath} 
\usepackage{amsmath}
\usepackage{amssymb}
\usepackage{dsfont}
\usepackage{makecell}
\usepackage{subcaption}
\usepackage[font=small,labelfont=bf]{caption}
\usepackage{tablefootnote}
\usepackage{soul}
\usepackage{enumitem}
\setlist{leftmargin=1em}

\usepackage[pagebackref=true,breaklinks=true,letterpaper=true,colorlinks,bookmarks=false]{hyperref}

\newcommand{\reducedstrut}{\vrule width 0pt height .9\ht\strutbox depth .9\dp\strutbox\relax}
\newcommand{\redbox}[1]{%
  \begingroup
  \setlength{\fboxsep}{0pt}%
  \colorbox[RGB]{255,216,213}{\reducedstrut#1\/}%
  \endgroup
}
\newcommand{\bluebox}[1]{%
  \begingroup
  \setlength{\fboxsep}{0pt}%
  \colorbox[RGB]{183,229,253}{\reducedstrut#1\/}%
  \endgroup
}
\newcommand{\greenbox}[1]{%
  \begingroup
  \setlength{\fboxsep}{0pt}%
  \colorbox[RGB]{224,240,208}{\reducedstrut#1\/}%
  \endgroup
}
\newcommand{\tablestyle}[2]{\setlength{\tabcolsep}{#1}\renewcommand{\arraystretch}{#2}\centering\footnotesize}
\NewCommandCopy{\sout}{\st}

\iccvfinalcopy 

\begin{document}

\title{DocTr: Document Transformer for \\ Structured Information Extraction in Documents}

\author{
Haofu Liao$^{1}$\thanks{Corresponding author \texttt{liahaofu@amazon.com}}
\quad Aruni RoyChowdhury$^{2\dag}$
\quad Weijian Li$^{3}$\thanks{Work done at AWS AI Labs}
\quad Ankan Bansal$^{1}$
\quad Yuting Zhang$^{1}$ \\
\quad Zhuowen Tu$^{1}$
\quad Ravi Kumar Satzoda$^{1}$
\quad R. Manmatha$^{1}$
\quad Vijay Mahadevan$^{1}$ \\
\\
$^{1}$AWS AI Labs \quad
$^{2}$MathWorks \quad
$^{3}$Amazon Physical Stores \quad
}

\maketitle
\ificcvfinal\thispagestyle{empty}\fi

\begin{abstract}
We present a new formulation for structured information extraction (SIE) from visually rich documents. It aims to address the limitations of existing IOB tagging or graph-based formulations, which are either overly reliant on the correct ordering of input text or struggle with decoding a complex graph. Instead, motivated by anchor-based object detectors in vision, we represent an entity as an anchor word and a bounding box, and represent entity linking as the association between anchor words. This is more robust to text ordering, and maintains a compact graph for entity linking. The formulation motivates us to introduce 1) a DOCument TRansformer (DocTr) that aims at detecting and associating entity bounding boxes in visually rich documents, and 2) a simple pre-training strategy that helps learn entity detection in the context of language. Evaluations on three SIE benchmarks show the effectiveness of the proposed formulation, and the overall approach outperforms existing solutions.
\end{abstract}

\section{Introduction}
\graphicspath{ {./images/} }

\begin{figure}[t]
  \centering
  \includegraphics[width=0.9\linewidth]{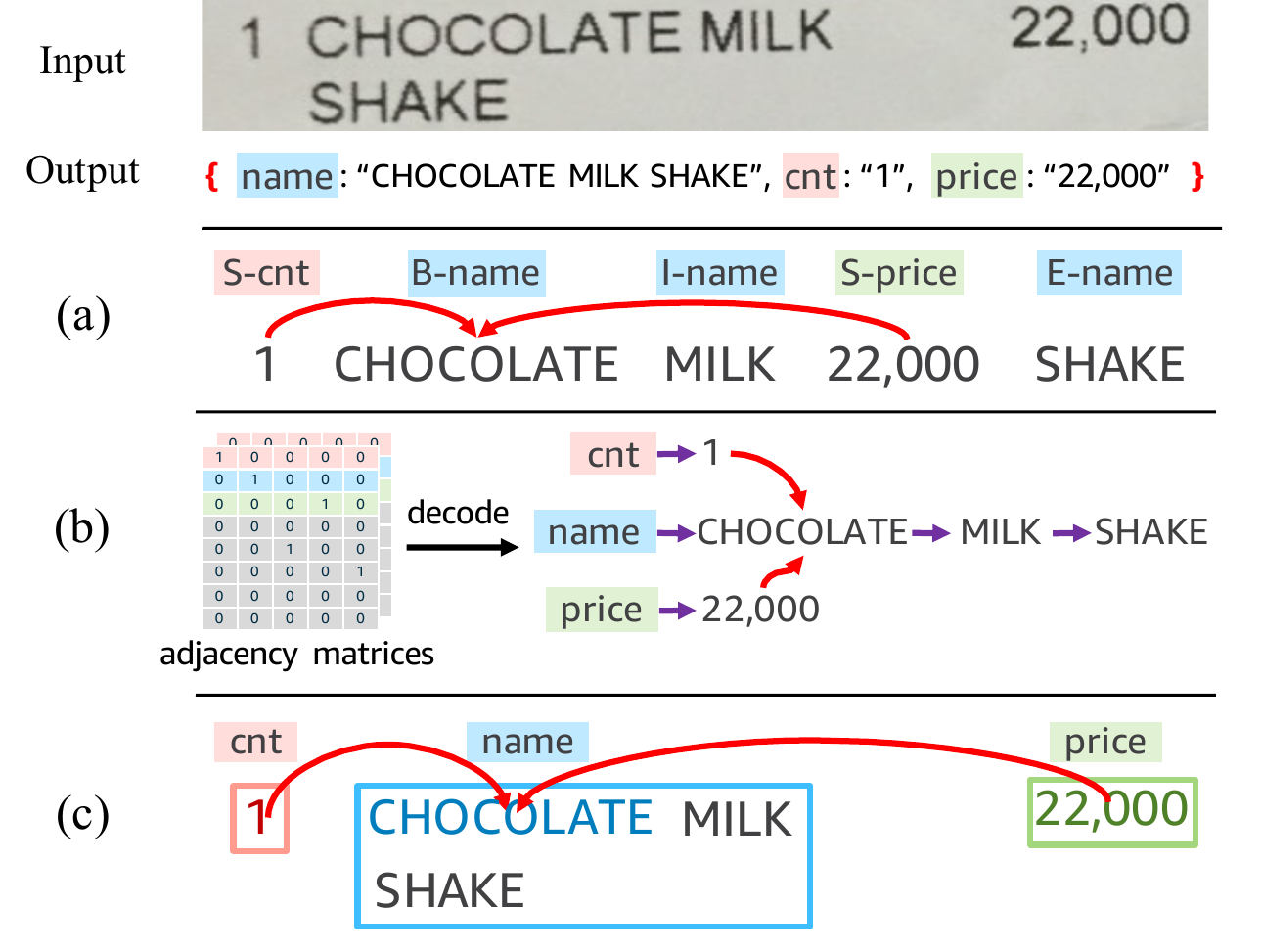}
  \caption{Structured information extraction problem formulations. Given an input image, 
  we aim to extract each entity (e.g., \bluebox{\texttt{name}}, \redbox{\texttt{count}}, or \greenbox{\texttt{price}}) and \textcolor[RGB]{225,50,35}{link} the related entities together. To address this task, \textbf{(a) IOB tagging}~\cite{ramshaw1999text} assigns a tag to each word to indicate if it is the beginning (B-), inside (I-), end (E-) of an entity or a single (S-) word entity. \textbf{(b) Graph based methods}~\cite{hwang2020spatial} take each word as a node, and use edges between words to indicate that the words belong to the same entity (\textcolor[RGB]{105,55,155}{purple} edges) or the underlying entities are linked (\textcolor[RGB]{225,50,35}{red} edges). A graph is generated by decoding from two adjacency matrices (one for each type of edges). \textbf{(c) Our formulation} represents an entity as an anchor word (colored words) and a box (colored bounding boxes), and represents entity linking via anchor word association (\textcolor[RGB]{225,50,35}{red} arrows).}
  \label{fig:intro_formulation}
  \vspace{-1.5em}
\end{figure}

Structured information extraction (SIE) from documents, as shown in Fig \ref{fig:intro_formulation}, is the process of extracting entities and their relationships, and returning them in a structured format. Structured information in a document is usually \textit{visually-rich} -- it is not only determined by the content of text but also the layout, typesetting, and/or figures and tables present in the document. Therefore, unlike the traditional information extraction task in nature language processing (NLP)~\cite{freitag2000machine,cardie1997empirical,riloff1996automatically} where the input is plain text (usually with a given reading order), SIE assumes the image representation of a document is available, and a pre-built optical character recognition (OCR) system may provide the unstructured text (i.e., without proper reading order). This is a practical assumption for day-to-day processing of business documents, where the documents are usually stored as images or PDFs, and the structured information, such as key-value pairs or line items (see Fig.~\ref{fig:conditioned_box}) from invoices and receipts, has been primarily obtained manually. This is time consuming and does not scale well. Hence, automating the document structured information extraction process with efficiency and accuracy is of great practical and scientific importance.

Structured information extraction is part of \textit{document intelligence}~\cite{cui2021document}, which focuses on the automatic reading, understanding, and analysis of documents. Early approaches to document intelligence usually address the problem purely from either a computer vision or an NLP perspective. The former takes the document as an image input and frames entity detection as object detection or instance segmentation~\cite{yang2017learning,schreiber2017deepdesrt}. The latter takes only the textual content of a document as the input, and addresses the problem with NLP solutions, such as IOB tagging via transformers~\cite{hwang2019post}.

Recently, models have also been proposed to pre-train on large-scale document collections and apply them to a wide variety of downstream document intelligence problems~\cite{xu2020layoutlm,hong2021bros,appalaraju2021docformer,lee2022formnet}. Such general-purpose models usually have the ability to make use of multi-modal inputs -- text from OCR, layout in the form of text locations, and visual features from images, and pre-training enables them to understand the basic structure of documents. Therefore, general-purpose models have demonstrated significant improvements on multiple document intelligence tasks, such as entity extraction~\cite{hong2021bros,lee2022formnet}, document image classification~\cite{xu2020layoutlm,appalaraju2021docformer}, and document visual question and answering~\cite{xu2021layoutlmv2,appalaraju2021docformer}.

For structured information extraction, existing general-purpose models rely on two broad approaches: 1) IOB tagging~\cite{ramshaw1999text} based methods~\cite{xu2020layoutlm,xu2021layoutlmv2,lee2022formnet}, and 2) graph based methods~\cite{hong2021bros,hwang2020spatial}. Both of these approaches suffer from inherent limitations. IOB tagging relies on the correct ``reading order'' or serialization of text, which however is not given by the OCR. As shown in Fig.~\ref{fig:intro_formulation}(a), the raster scan order of OCR text separates \texttt{I-name} and \texttt{E-name}. When there are multiple \texttt{name} entities, it could be non-trivial to know which \texttt{I-name}/\texttt{E-name} word belongs to which \texttt{name} entity. Graph-based methods (Fig.~\ref{fig:intro_formulation} (b)) can result in complex graphs with many words in a document (i.e., many nodes in the graph). Therefore, decoding the entities and their relationships from the adjacency matrices is error-prone.

Given the limitations of existing work, we make the following contributions in this paper:
\vspace{-0.4em}
\begin{itemize}
\setlength\itemsep{0.1em}
\item We introduce \textit{a new formulation} for SIE where we represent an entity as an anchor word along with a box, and regard the problem as an anchor word based entity detection and association problem (Fig~\ref{fig:intro_formulation} (c)). Thus, we extract entities via bounding boxes and do not depend on the reading order of input. We assign each entity with an anchor word, resulting in a compact graph of entity relations (e.g., the anchor word links in Fig~\ref{fig:intro_formulation} (c)), which facilitates decoding structured information.

\item We develop \textit{a new model}, called DOCument TRansformer (DocTr), which combines a language model and visual object detector for joint vision-language document understanding. We note that the recognition of an anchor word is largely a language-dependent task, while the detection of entity boxes is a more vision-dependent task. Therefore, DocTr is an intuitive approach to target this problem under the proposed formulation.

\item We propose \textit{a new pre-training task}, called masked detection modeling (MDM), that matches our formulation and helps learn box prediction in the context of language. Our experimental results show that 1) the proposed formulation addresses SIE better than IOB tagging or graph-based solutions, 2) MDM is a more effective pre-training task, in particular when worked together with the new formulation,  and 3) the overall approach outperforms existing solutions on three SIE tasks.
\end{itemize}

\section{Related Work}

\textbf{General-purpose document understanding.}
General-purpose approaches aim to develop a backbone model for document understanding, which is then adapted to address downstream document understanding tasks. LayoutLM~\cite{xu2020layoutlm,xu2021layoutlmv2,huang2022layoutlmv3} is an early approach that pre-trains on a large-scale document dataset. It introduces masked vision-language modeling and layout information for document understanding pre-training. BROS~\cite{hong2021bros} improves LayoutLM via better encoding of the spatial information and introducing a pre-training loss for understanding text blocks in 2D. DocFormer~\cite{appalaraju2021docformer} introduces a new architecture and pre-training losses to better leverage text, vision and spatial information in an end-to-end fashion. FormNet~\cite{lee2022formnet} encodes neighborhood context for each token using graph convolutions and introduces an attention mechanism to address imperfect serialization. StrucText~\cite{li2021structext} proposes to extract multi-modal semantic features at both token level, word-segment level and/or entity level. Donut~\cite{kim2022ocr} proposes an OCR free solution that is pre-trained to predict document text from images. It is an encoder-decoder model that can directly decode the expected outputs as text for downstream tasks.

\textbf{Structured information extraction (SIE).}
Early approaches \cite{yang2017learning,katti2018chargrid,denk2019bertgrid} formulate the SIE problem as a computer vision problem to either segment or detect entities from documents. However, they cannot address linking of entities due to the limitation of the formulation. With the advent of transformers~\cite{vaswani2017attention} and their success in NLP, more recent approaches~\cite{liu2019roberta,zhang2020trie,garncarek2021lambert} address SIE by incorporating layout/visual information with text inputs to transformers, and extract entities via a NLP formulation~\cite{ramshaw1999text}. Other approaches~\cite{liu2019graph,wei2020robust,yu2021pick} propose to regard the text inputs as the nodes in a graph and model the relationship of text inputs via graph neural networks. To extract the relationship between entities, SPADE~\cite{hwang2020spatial} introduces a graph decoding scheme on learned pairwise affinities between extracted entities.

\textbf{Table detection and recognition (TDR).}
TDR is the task of detecting and recognizing tabular structures from document images. Both SIE and TDR focus on returning information in a structured way from documents. However, unlike SIE where the spatial relationship of entities are unconstrained, TDR assumes a tabular structure of entities (i.e., table cells) and leverages this prior knowledge in the model design and post-processing~\cite{prasad2020cascadetabnet,zheng2021global,nassar2022tableformer}. Moreover, SIE requires returning a semantic label for each entity which demands an understanding of the text, while TDR does not distinguish between types of table cells but focuses more on table layout. Therefore, the existing approaches~\cite{prasad2020cascadetabnet,zheng2021global,nassar2022tableformer} to TDR are vision-only approaches.

\textbf{TextVQA.}
Given an input image, TextVQA aims to answer questions related to the text in image. Similar to SIE, existing TextVQA approaches~\cite{singh2019towards,hu2020iterative,gao2021structured,biten2022latr} employ multi-modal models that take both the OCR and image as inputs. However, for TextVQA, the answers are typically single entities. It can be challenging to address the problem with TextVQA if we aim to return multiple entities in a structured way, and if an image could have multiple of such structures.

\textbf{Scene graph generation (SGG).}
Generating scene graphs can be regarded as a form of SIE for natural images. SGG methods~\cite{johnson2015image,xu2017scene,zellers2018neural,yang2018graph,tang2020unbiased} detect objects as the nodes of scene graphs, and construct edges of scene graphs by identifying the pairwise relationships between objects. This is similar to our formulation of SIE where we extract entities via anchor word guided object detection, and link entities by learning to output their pairwise affinities. 
\section{Approach}
\graphicspath{ {./images/} }

\subsection{Structured Information Extraction} \label{sec:new_formulation}

\begin{figure}
  \centering
  \includegraphics[width=\linewidth]{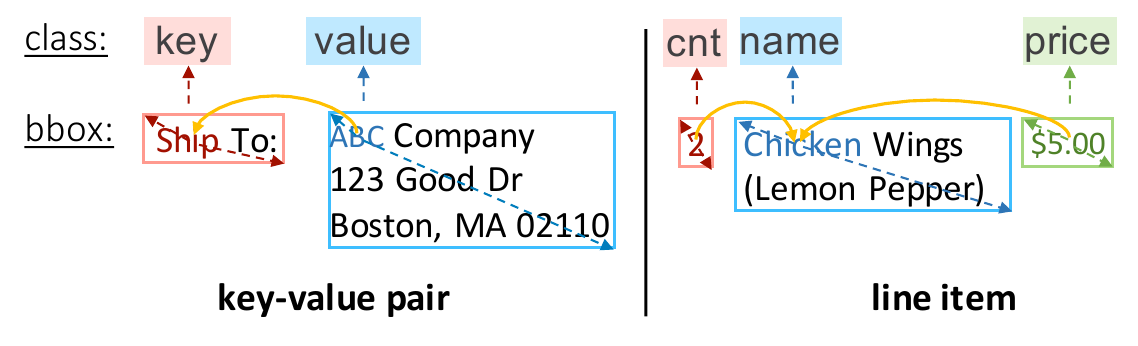}
  \vspace{-1.5em}
  \caption{Illustration of our formulation for structured information extraction of two types of relations. We first identify the ``anchor words'' of entities (which are the first words of entities in this example, e.g., ``Ship'' or ``Chicken''). Then, from each anchor word, we extract the entity by predicting (dotted arrows) its class label and bounding box. We link entities by linking (yellow arrows) their anchor words. For key-value pair relation, we link ``Ship'' to ``ABC''. For line item relation, we link ``2'' and ``\$5.00'' to ``Chicken''.}
  \label{fig:conditioned_box} \vspace{-1.2em}
\end{figure}

\paragraph{Problem Formulation.} Following prior work, we assume the input is the image of a document page, and a pre-built OCR system is applied to detect and recognize the words. The goal of a structured information extraction system for document understanding is to extract a set of grouped entities $\mathcal{G}=\{\mathbf{G}_i\}$, where each entity group $\mathbf{G}_i=\{\mathbf{e}_{ij}\}$ is a set of entities with predefined relations. As shown in Fig.~\ref{fig:conditioned_box}, an entity group may be a key and value pair, or a line item containing the name, count and price entities. We denote an entity as $\mathbf{e}=(t,c,b)$ where $t$, $c$ and $b$ are the text, class label, and location (bounding box) of the entity, respectively. Note that, with OCR inputs, this formulation of an entity can be reduced to $\mathbf{e}=(c,b)$, because the text $t$ can be obtained by aggregating the OCR text inside $b$.

Next, we propose a \textit{new formulation} to address structured information extraction. We propose to address \textit{entity extraction} via anchor word guided detection and \textit{entity linking} via anchor word association. The former extracts entities $\{\mathbf{e}_i\}$, and the latter links entities into groups $\{\mathbf{G}_i\}$.

\vspace{-1em}
\paragraph{Entity Extraction via Anchor Word Guided Detection.} To extract an entity $\mathbf{e}$, we first introduce a new concept called \textit{anchor word}, which is a designated word of an entity. In Fig.~\ref{fig:conditioned_box}, we select the first word of an entity as the anchor word, e.g.,  ``ABC'' is the anchor word for \texttt{value}, and ``Chicken'' is the anchor word for \texttt{name}. Other designations of anchor words are possible (see Sec.~\ref{sec:alation_study}). An anchor word may be regarded as the representation of an entity. Since the goal of extracting an entity $\mathbf{e}=(c,b)$ is to find its class label $c$ and bounding box $b$, they may then be represented by an anchor word. As shown in Fig.~\ref{fig:conditioned_box}, we associate each anchor with a label and a bounding box. For example, the anchor word ``Ship'' is associated with a label \texttt{key} and a bounding box that encloses the entity ``Ship To:''. Therefore, the task of extracting an entity may be seen as first identifying its anchor word, and then obtaining the label and bounding box associated with it.

\vspace{-1em}
\paragraph{Entity Linking via Anchor Word Association.} 
We define an entity group as consisting of a \textit{primary} entity, and all the other entities in the group are \textit{secondary}. The anchor word of a primary/secondary entity is the primary/secondary anchor word. Once anchor words have been identified, linking entities into entity groups is equivalent to associating anchor words. To establish such association, we first select the primary anchor words of entity groups, and then all the secondary anchor words from the same group are linked to the primary anchor word. The definition of a primary entity may vary. For key-value pairs, the primary anchor words may simply be those anchor words labeled as \texttt{key}. For more general entity groups, we designate a primary anchor word based on the task/data. For example, we choose \texttt{name}'s anchor word ``Chicken'' as the primary anchor in Fig.~\ref{fig:conditioned_box}. Other ways of choosing primary anchors are possible (See Sec.~\ref{sec:alation_study}). 
Links between primary and secondary anchor words are represented by a binary matrix $\mathbf{M} \in \{0, 1\}^{m \times n}$. $\mathbf{M}_{ij}=1$ indicates that the $i$th primary anchor word, and $j$th secondary anchor word are linked. Otherwise, $\mathbf{M}_{ij}=0$.

\subsection{DocTr: Document Transformer} \label{sec:doctr}

\begin{figure}
  \centering
  \vspace{-0.5em}
  \includegraphics[width=\linewidth]{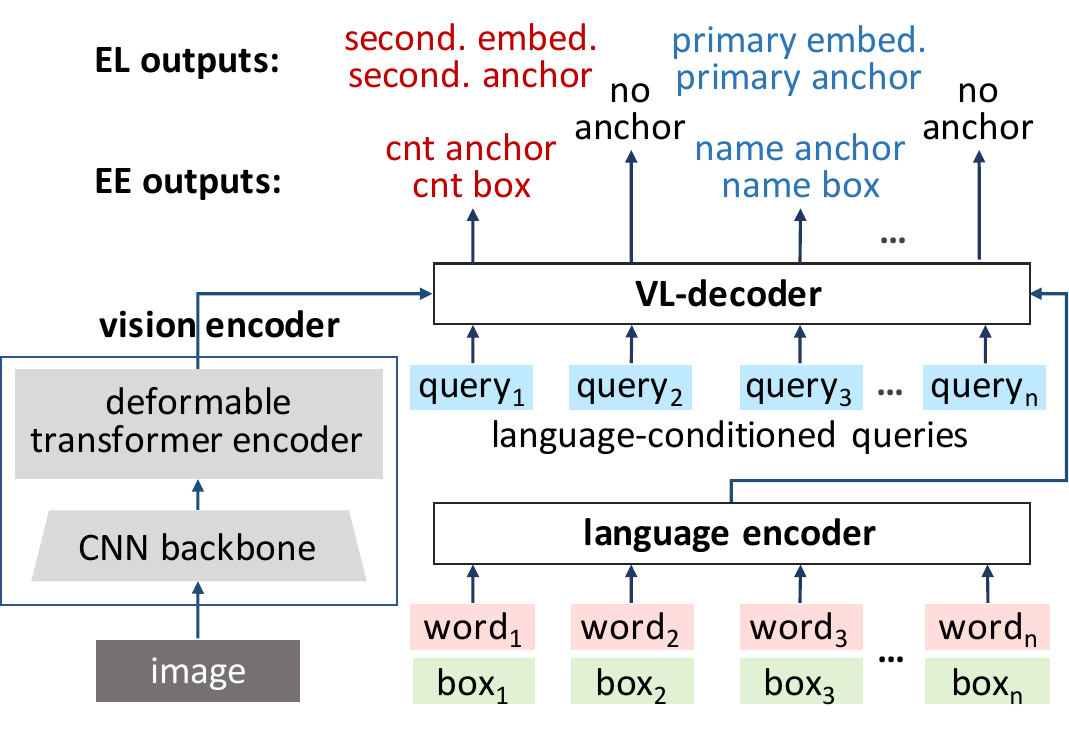}
  \vspace{-1em}
  \caption{Overall architecture of DocTr. The vision encoder extracts visual features from a document image. The language encoder extracts language features from OCR text and bounding boxes (i.e., document layout information). The VL-decoder uses language-conditioned queries to decode structured information from visual and language features. For entity extraction (EE), each query decodes an anchor word label and an entity box. For entity linking (EL), each query decodes an association embedding of an anchor word and a primary/secondary anchor label.}
  \label{fig:doctr} \vspace{-1.2em}
\end{figure}

DocTr is a multi-modal transformer that takes both the document image and OCR words (text and position) as input. Unlike existing encoder-only approaches~\cite{xu2021layoutlmv2,appalaraju2021docformer,li2021structext}, DocTr has an encoder-decoder architecture with 1) two dedicated encoders to encode vision and language features separately, and 2) a vision-language decoder to decode anchor word based outputs for entity extraction and entity linking. An overview of the DocTr architecture is shown in Fig.~\ref{fig:doctr}.

\vspace{-1em}
\paragraph{Vision Encoder.}
The vision encoder is adapted from Deformable DETR~\cite{zhu2020deformable}. It consists of a CNN backbone with multi-scale visual feature extraction, and a deformable transformer encoder for efficient encoding of visual features. Compared with vanilla transformer based vision encoders, this design is more lightweight due to the use of deformable attention, which has linear complexity with respect to the spatial size of image feature maps instead of the quadratic complexity using standard self-attention. As a result, it is capable of encoding high-resolution multi-scale visual features for better detection of small objects/entities. 

This vision encoder is shown to work effectively with a transformer decoder for end-to-end object detection~\cite{carion2020end}. This is helpful to our formulation of entity extraction, where we convert this task into an anchor word guided object detection problem. We also highlight the differences from existing encoder-only methods where the visual features -- either region-based~\cite{xu2020layoutlm,li2021structext} or grid-based~\cite{xu2021layoutlmv2,appalaraju2021docformer} -- are extracted with a pre-trained CNN model; they are sent to the transformer encoder along with OCR inputs without dedicated network components for decoding entity bounding boxes.

\vspace{-1em}
\paragraph{Language Encoder.}
The language encoder is a transformer model adapted from the BERT architecture~\cite{devlin2019bert}. We follow LayoutLM~\cite{xu2020layoutlm} to include the layout information (i.e., 2D position embeddings of OCR) along with the OCR text as input. However, no visual information is added since it has already been addressed by the vision encoder. The language encoder is critical to our formulation for the identification of anchor words, which is a language-dependent task. 

\vspace{-1em}
\paragraph{Vision-Language Decoder with Language-Conditioned Queries.} The architecture of the vision-language decoder is similar to the decoder of the Deformable DETR transformer model~\cite{zhu2020deformable} - with two major differences to facilitate the decoding of vision-language inputs. Each decoder layer has two cross-attention modules to decode from vision and language inputs respectively. For vision, we apply \textit{deformable cross-attention} (similar to Deformable DETR) to efficiently decode from high-resolution visual features. For language, we apply \textit{language-conditioned cross-attention} to decode from the discrete OCR language features. 

Specifically, we introduce \textit{language-conditioned queries} to better leverage the OCR inputs and obviate the need for bipartite matching between predicted and ground truth entities. The original DETR-like decoder queries~\cite{carion2020end,zhu2020deformable} do not have explicit semantic meanings at the beginning. Hence, DETR requires finding the most plausible matching between a prediction and ground truth, which is less effective and impedes the training. For document understanding with OCR inputs, we consider a one-to-one mapping between OCR inputs and decoder queries. That is, we have the same number of queries as the number of OCR inputs to the language encoder, and the $i$th query is mapped to the $i$th OCR input (see Fig.~\ref{fig:doctr}). This mapping can be simply modeled as cross-attention between queries and language embeddings by using the same position embedding for both inputs. Let $\mathbf{Q} \in \mathbb{R}^{L \times d}$ be a set of $L$ decoder queries each with dimension $d$ (packed as a matrix), $\mathbf{V} \in \mathbb{R}^{L \times d}$ be the set of output embeddings from the language encoder, and $\mathbf{P} \in \mathbb{R}^{L \times d}$ be a set of position embeddings. Then, the cross attention with language-conditioned queries can be written as:
\begin{equation} \label{eq:language_conditioined_ca}
\resizebox{0.9\columnwidth}{!}{%
$\text{CrossAttn}(\mathbf{Q}, \mathbf{V}, \mathbf{P}) = \text{softmax}(\frac{(\mathbf{Q}+\mathbf{P})(\mathbf{V}+\mathbf{P})^T}{\sqrt{d}})\mathbf{V},$}
\end{equation}
where $\sqrt{d}$ is a scaling factor~\cite{vaswani2017attention}. 
This mapping assigns each query with an explicit linguistic semantic meaning -- the $i$-th decoder output now corresponds to the $i$-th input text token, via the $i$-th decoder query. Thus, we can directly match entities with queries without the bipartite matching required by the default DETR decoder formulation~\cite{carion2020end,zhu2020deformable}.

\vspace{-1em}
\paragraph{Entity Extraction and Linking Outputs.} The decoder has two sets of outputs for entity extraction and entity linking respectively (see Fig.~\ref{fig:doctr}). For entity extraction, each output is a class label and a bounding box which uniquely decide an entity. Because each query (and thus its corresponding output) is mapped to an OCR input, the class label indicates whether the underlying OCR input is an anchor word, and the type of entity it represents. For entity linking, each output is a binary class label and an embedding vector. The binary class label indicates whether the OCR input is a primary anchor word. The embedding vector is for the linking of anchor words, and we use different embeddings for primary and secondary anchor words. Let $\mathbf{E}_{\text{p}} \in \mathbb{R}^{m \times h}$ be a set of $m$ primary  embeddings, and $\mathbf{E}_{\text{s}} \in \mathbb{R}^{n \times h}$ be a set of $n$ secondary embeddings, the predicted affinity matrix for entity linking is computed as $\mathbf{\hat{M}} = \text{sigmoid}(\mathbf{E}_{\text{p}} \mathbf{E}_{\text{s}}^T), \mathbf{\hat{M}}\in (0,1)^{m \times n}$.

\subsection{Architecture Details} \label{sec:architecture}

For the vision encoder, we use a ResNet50 backbone and a 6-layer deformable transformer encoder~\cite{zhu2020deformable}. The backbone is initialized with ImageNet pretrained weights, and outputs three scales of visual features. The multi-scale visual features are transformed into a sequence with 2D ``sine'' position embeddings before sending to the deformable transformer encoder. For the language encoder, we use a 12-layer transformer encoder with the same architecture settings as the BERT-base model~\cite{devlin2019bert}. In addition to BERT's text embeddings and 1D position embeddings, we also add 2D position embeddings~\cite{xu2020layoutlm} to include layout information of the document as the input. The 2D position embeddings are learned embeddings with random initialization. The VL-decoder has 6 layers, where each layer consists of a self-attention module, a deformable cross-attention module~\cite{zhu2020deformable} and a standard cross-attention module~\cite{vaswani2017attention} (see supplementary material for detailed architecture of VL-decoder layers).

\subsection{Training and Pre-training} \label{sec:training}

\begin{figure}
  \centering
  \includegraphics[width=1.0\linewidth]{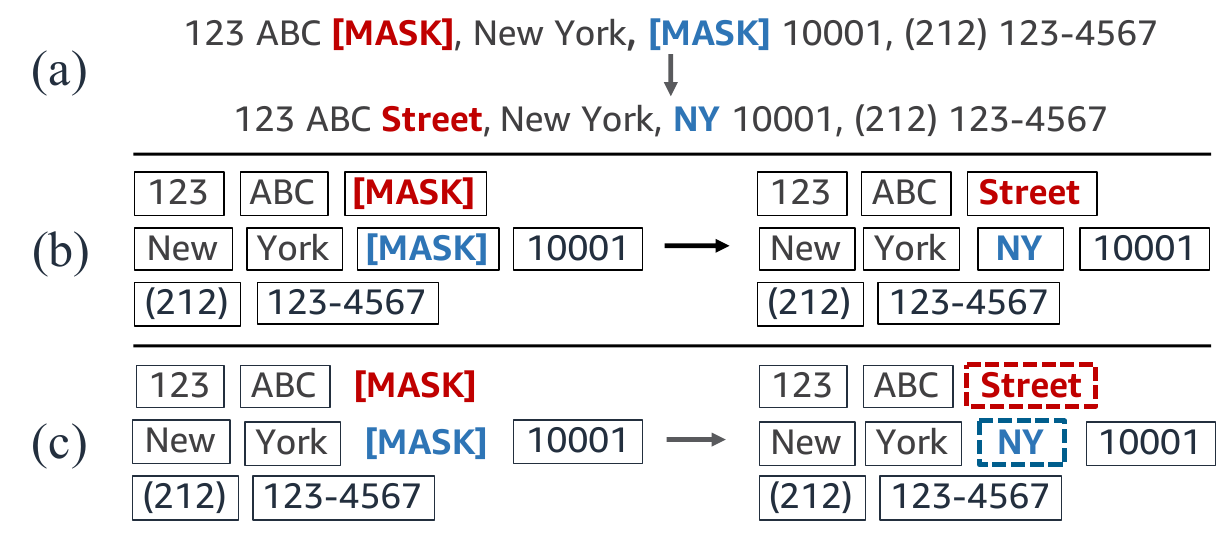}
  \captionof{figure}{Illustration of of masked detection modeling (MDM) and comparison with masked language modeling (MLM)~\cite{devlin2019bert} and masked vision-language modeling (MVLM)~\cite{xu2020layoutlm}. \textbf{(a) MLM} takes only text as input, and requires predicting the masked text input. \textbf{(b) MVLM} takes both OCR texts and boxes as input. But only the text part is masked and to be predicted. \textbf{(c) MDM (ours)} takes one step further by masking both the texts and boxes, and requires predicting the masked boxes and their corresponding texts.}
  \label{fig:pretraining_comparison} \vspace{-1em}
\end{figure}

\paragraph{Entity Extraction and Linking Objectives.} The entity extraction objective is similar to the one used in DETR~\cite{carion2020end} except that we do not need the bipartite matching due to the use of language-conditioned queries (as introduced in Sec.~\ref{sec:doctr}). Specifically, given a set of $N$ OCR inputs, the language-conditioned queries yields $N$ entity extraction outputs $\mathbf{\hat{E}}= \{\mathbf{\hat{e}}_i\}_{i=1}^N$. For a document with $M$ entities, we also construct a ground truth $\mathbf{E}= \{\mathbf{e}_i\}_{i=1}^N$ of size $N$. Here, $\mathbf{\hat{e}}_i$ and $\mathbf{e}_i$ denote the predicted and ground truth entities of the $i$th OCR input, respectively. Note that not every OCR word is an anchor word, and thus it may have no associated entity. In this case, we say that the ground truth of the input OCR is an empty entity, i.e., $\mathbf{e}=\varnothing$, and there are in total $N-M$ empty entities in $\mathbf{E}$. If we denote a non-empty entity as $\mathbf{e}=(c,b)$ and a predicted entity as $\mathbf{\hat{e}}=(\hat{p}, \hat{b})$, where $c$ is the ground truth entity label, $\hat{p}$ is the predicted entity label probability, and $b$/$\hat{b}$ is the ground truth/predicted bounding box, then we write the entity extraction loss as
\begin{equation} \label{eq:entity_extraction}
    \resizebox{0.91\columnwidth}{!}{%
    $\mathcal{L}_{\text{EE}}(\mathbf{E}, \mathbf{\hat{E}}) = \sum_i[-\log{\hat{p}_i(c_i)} + \lambda \mathds{1}_{\{\mathbf{e}_i \neq \varnothing\}}\mathcal{L}_{\text{bbox}}(b_i, \hat{b}_i)],$
    }
\end{equation}
where $\hat{p}_i(c_i)$ is the predicted probability of entity being labeled as $c_i$, $\mathcal{L}_{\text{bbox}}$ is a bounding box loss~\cite{carion2020end}, and $\mathds{1}_{\{\mathbf{e}_i \neq \varnothing\}}$ means we only compute $\mathcal{L}_{\text{bbox}}$ for non-empty entities.

The entity linking loss consists of two parts, primary anchor classification and linking classification. Let $\mathbf{\hat{L}}$ be a set of primary anchor classification outputs and $\mathbf{L}$ be its binary ground truth labels. Let $\mathbf{\hat{M}}$ and $\mathbf{M}$ be the predicted and ground truth entity linking affinity matrices, respectively. Then, we can simply write the entity linking loss as
\begin{equation} \label{eq:entity_linking}
    \mathcal{L}_{\text{EL}}(\mathbf{L}, \mathbf{\hat{L}}, \mathbf{M}, \mathbf{\hat{M}}) = \text{BCE}(\mathbf{L}, \mathbf{\hat{L}}) + \beta \text{BCE}(\mathbf{M}, \mathbf{\hat{M}}),
\end{equation}
where $\text{BCE}$ denotes the binary cross-entropy loss.

\vspace{-1em}
\paragraph{Pre-training.} We pre-train DocTr on a large-scale dataset of unlabeled document images. For simplicity of modeling, we only include one pre-training task, termed \textit{masked detection modeling (MDM)}, for DocTr which we find sufficient for downstream tasks. Since pre-training is not the main focus of this work, we leave the exploration of other pre-training strategies~\cite{xu2021layoutlmv2,hong2021bros,appalaraju2021docformer} for future work. Fig.~\ref{fig:pretraining_comparison} illustrates MDM and compares it with related pre-training tasks. MDM is an extension of \textit{masked vision-language modeling (MVLM)}~\cite{xu2020layoutlm,xu2021layoutlmv2}. Both MDM and MVLM take OCR text and boxes as input. However, MVLM only randomly masks the text inputs. Instead, MDM randomly masks both the text inputs and their boxes. Specifically, we replace text with \texttt{[MASK]} and set boxes to \texttt{[0,\,0,\,0,\,0]}. Then, we train DocTr to predict both the masked texts and their corresponding boxes. Note that this task is similar to object detection. Thus, the objective function can be written in the same way as Eq.~(\ref{eq:entity_extraction}), where the first term is for masked text classification, and the second term is for masked box regression. Also note that for MDM, the input image is not masked so that a model can better learn how to leverage the visual information to locate and identify the masked inputs.
\section{Experiments}
\graphicspath{ {./images/} }

\paragraph{Datasets and Tasks.} We use three datasets in our experiments, IIT-CDIP document collection~\cite{lewis2006building}, CORD~\cite{park2019cord} and FUNSD~\cite{jaume2019funsd}. We follow the convention in the literature~\cite{xu2020layoutlm,xu2021layoutlmv2,hong2021bros,appalaraju2021docformer} to pre-train DocTr on the IIT-CDIP document collection, which is a large-scale dataset with 11 million unlabeled documents.  CORD~\cite{park2019cord} is a receipt dataset with 800 training, 100 validation, and 100 testing samples. Each receipt in this dataset is labeled with a list of line items and key-value pair groups. FUNSD~\cite{jaume2019funsd} consists of scanned forms, with 149 training and 50 testing examples. Each form is labeled with key/value entities together with links to indicate which keys and values are associated.

\begin{figure}[t]
  \centering
  \includegraphics[width=1.0\linewidth]{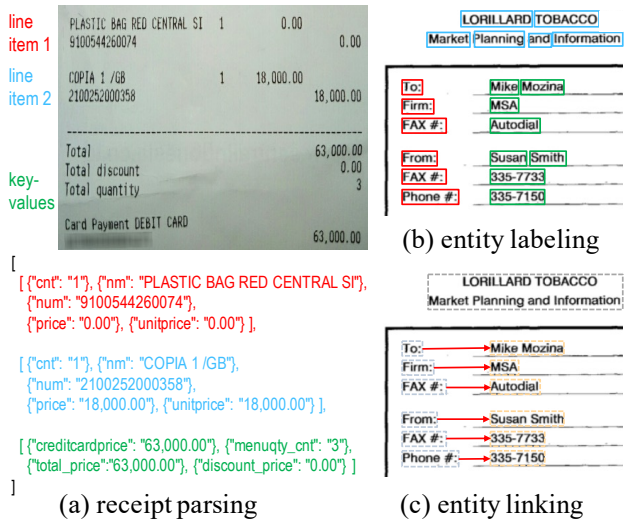}
  \vspace{-1.5em}
  \captionof{figure}{Illustration of the three tasks in the experiments.}
  \label{fig:tasks}
  \vspace{-1.5em}
\end{figure}

We evaluate our model's performance on three tasks, receipts parsing, entity labeling and entity linking. For \textit{receipt parsing}, a model not only has to extract each receipt's entities but also correctly link entities to form line items and key-value pair groups. Fig.~\ref{fig:tasks} (a) shows a sample receipt from CORD and its expected output after parsing. The sample contains two line items and four key-value pairs. For line items, it requires identifying each line item related entity (class and text) and group the entities of the same line item together. For key-value pairs, we identify class labels of the keys and return only text of the corresponding values. We use the same evaluation protocols and metrics as defined in \cite{hwang2020spatial} to evaluate the receipt parsing performance.

\textit{Entity labeling} and \textit{entity linking} are commonly adopted tasks~\cite{xu2020layoutlm,hwang2020spatial} to evaluate a pre-trained model's performance, which however are simplified versions of what we have defined in Sec.~\ref{sec:new_formulation}. Entity labeling requires assigning a class label to each word of the document. Fig.~\ref{fig:tasks} (b) shows a sample from FUNSD where the task is to identify if a word belongs to a key (red), a value (green) or a title (blue). In entity linking, the assumption is that the key/value entities are correctly detected, and the task is to identify which keys and values should be linked (See Fig.~\ref{fig:tasks} (c), red arrows). We evaluate entity labeling/linking by checking if the words/links are correctly labeled using F1-score as the metric.

\subsection{Comparison with Existing Solutions}

We compare DocTr with the existing methods on receipts parsing, entity labeling, and entity linking tasks, respectively.

For \textit{receipts parsing}, SPADE~\cite{hwang2020spatial} and Donut~\cite{kim2022ocr} are the only two other publicly available solutions (to the best of our knowledge) that address this task on CORD. The other existing general-purpose models~\cite{xu2021layoutlmv2,hong2021bros, huang2022layoutlmv3} are not able to directly address this structured information extraction task out-of-the-box. For a fair comparison with our method, we fine-tune the officially released general-purpose models under two settings: using the standard IOB tagging for receipts parsing or using our proposed formulation. From Table~\ref{tab:receipts_parsing}, we can see that DocTr outperforms general-purpose models BROS, LayoutLMv2 and LayoutLMv3 by a noticeable margin when they are fine-tuned with the IOB tagging setting. When fine-tuned with our proposed formulation, the general-purpose models' performance improved but they are still behind DocTr, which shows the effectiveness of the proposed encoder-decoder solution for the anchor word based structure information extraction.

For \textit{entity labeling}, we follow the general-purpose models~\cite{xu2020layoutlm} to only fine-tune DocTr for IOB tagging and evaluate based on its IOB tagging outputs. We note that this is less favorable for DocTr since the architecture and is dedicated to address our new formulation, and the pre-training strategy is not a main focus of this paper. However, we observe DocTr noticeably outperforming the existing solutions with comparable model sizes (``Base'' models in Table~\ref{tab:entity_labeling}). Even when compared with larger pre-trained models, DocTr's performance is comparable or better on the CORD dataset. 

For \textit{entity linking}, we apply the objective introduced in Eq.~(\ref{eq:entity_linking}) to train our model to link keys and values in FUNSD documents. We remove the entity extraction loss (Eq.~(\ref{eq:entity_extraction})) but use ground truth entities as per the task definition. The results are shown in Table~\ref{tab:entity_linking} -- DocTr also outperforms the existing solutions by a noticeable margin in this task.

\begin{table}[t]
\captionsetup{font=footnotesize}
\begin{minipage}{0.45\columnwidth}{
    \begin{center}
        \tablestyle{4pt}{1.05}
        \begin{tabular}{lc}
        \bf model & \bf F1 \\
        \midrule
        $\textrm{Donut}$~\cite{kim2022ocr}\textsuperscript{$\dagger$} & 87.8 \\
        $\textrm{SPADE}$~\cite{hwang2020spatial} & 92.5 \\
        \midrule
        $\textrm{LayoutLMv2}$~\cite{xu2021layoutlmv2} w/ IOB & 91.4 \\
        $\textrm{BROS}$~\cite{hong2021bros}  w/ IOB & 91.8 \\
        $\textrm{LayoutLMv3}$~\cite{huang2022layoutlmv3}  w/ IOB & 92.2 \\
        \midrule
        $\textrm{LayoutLMv2}$~\cite{xu2021layoutlmv2} w/ ours & 92.7 \\
        $\textrm{BROS}$~\cite{hong2021bros} w/ ours & 92.9 \\
        $\textrm{LayoutLMv3}$~\cite{huang2022layoutlmv3} w/ ours & 93.6 \\
        \midrule
        $\textrm{\bf DocTr (ours)}$ & \bf{94.4} \\
        \end{tabular}
    \vspace{-.5em}
    \caption{Comparison with existing solutions on receipts parsing with the CORD dataset.}
    \label{tab:receipts_parsing}
    \end{center}
}\end{minipage}
\hspace{1em}
\begin{minipage}{0.45\columnwidth}{
    \begin{center}
        \tablestyle{4pt}{1.05}
        \begin{tabular}{lc}
            \multicolumn{1}{c}{\bf model} & \bf F1 \\
            \midrule
            $\textrm{SPADE}$~\cite{hwang2020spatial}                  & 41.7  \\
            $\textrm{BROS}$~\cite{hong2021bros}                       & 71.5  \\
            $\textrm{StructText}$~\cite{li2021structext}              & 44.1  \\
            $\textrm{\bf DocTr (ours)}$                               & \bf 73.9  \\

        \end{tabular}
    \vspace{-.5em}
    \caption{Comparison with existing solutions on entity linking with the FUNSD dataset.}
    \label{tab:entity_linking}
    \end{center}
}\end{minipage}

\vspace{.2em}
\scriptsize{\textsuperscript{$\dagger$}We take the official model from~\cite{kim2022ocr} and report numbers using the metric from~\cite{hwang2020spatial}.}
\vspace{-1em}
\end{table}

\begin{table}[t]
    \centering
    \tablestyle{4pt}{1.05}
    \resizebox{\linewidth}{!}{\begin{tabular}{lccc}
        \multicolumn{1}{c}{\bf model} & \bf FUNSD & \bf CORD & \bf \#params  \\
        \midrule
        $\textrm{SPADE}$~\cite{hwang2020spatial}                  & 71.6                  & -                     & -    \\
        $\textrm{LayoutLM}_{\rm BASE}$~\cite{xu2020layoutlm}      & 78.7                  & 94.7                  & 113M \\
        $\textrm{BROS}_{\rm BASE}$~\cite{hong2021bros}            & 83.1                  & 96.5                  & 110M \\
        $\textrm{DocFormer}_{\rm BASE}$~\cite{hong2021bros}       & 83.3                  & 96.3                  & 183M \\
        $\textrm{LayoutLMv2}_{\rm BASE}$~\cite{xu2021layoutlmv2}  & 82.8                  & 95.0                  & 200M \\
        $\textrm{StructText}$~\cite{li2021structext}              & 83.4                  & -                     & 107M \\
        $\textrm{\bf DocTr} (ours)$                               & \bf{84.0}      & \bf 98.2              & 153M \\
        \midrule
        $\textrm{LayoutLM}_{\rm LARGE}$~\cite{xu2020layoutlm}     & 79.0                  & 95.0             & 343M  \\
        $\textrm{BROS}_{\rm LARGE}$~\cite{hong2021bros}           & 84.5                  & \bf 97.3             & 340M  \\
        $\textrm{DocFormer}_{\rm LARGE}$~\cite{hong2021bros}      & 84.5                  & 97.0             & 536M  \\
        $\textrm{LayoutLMv2}_{\rm LARGE}$~\cite{xu2021layoutlmv2} & 84.2                  & 96.0             & 426M  \\
         $\textrm{FormNet}$~\cite{lee2022formnet}                 & \bf{84.7}      & \bf{97.3} & 345M \\
    \end{tabular}}
    \vspace{-0.7em}
    \caption{Comparison with existing solutions on entity labeling (with FUNSD and CORD datasets).}
    \label{tab:entity_labeling} \vspace{-1em}
\end{table}

\begin{table}[!t]
    \centering
    \tablestyle{4pt}{1.05}
    \begin{tabular}{ccc}
    \bf formulation                        & \bf text serial. & \bf parsing (C) \\
    \midrule
    IOB tagging~\cite{ramshaw1999text}   & raster scan      & 93.2  \\
    SPADE~\cite{hwang2020spatial}        & raster scan      & 93.0  \\
    \bf DocTr (ours)                     & raster scan      & \bf 94.4  \\
    \cmidrule{1-3}
    IOB tagging~\cite{ramshaw1999text}   & oracle           & 94.1  \\
    SPADE~\cite{hwang2020spatial}        & oracle           & 93.9  \\
    \bf DocTr (ours)                     & oracle           & \bf 95.0  \\
    \end{tabular}
    \vspace{-0.7em}
    \caption{Comparison of different SIE formulations under two text serialization settings, raster scan and oracle.}
    \label{tab:formulation} \vspace{-1em}
\end{table}

\begin{figure}[t]
  \centering
  \includegraphics[width=\linewidth]{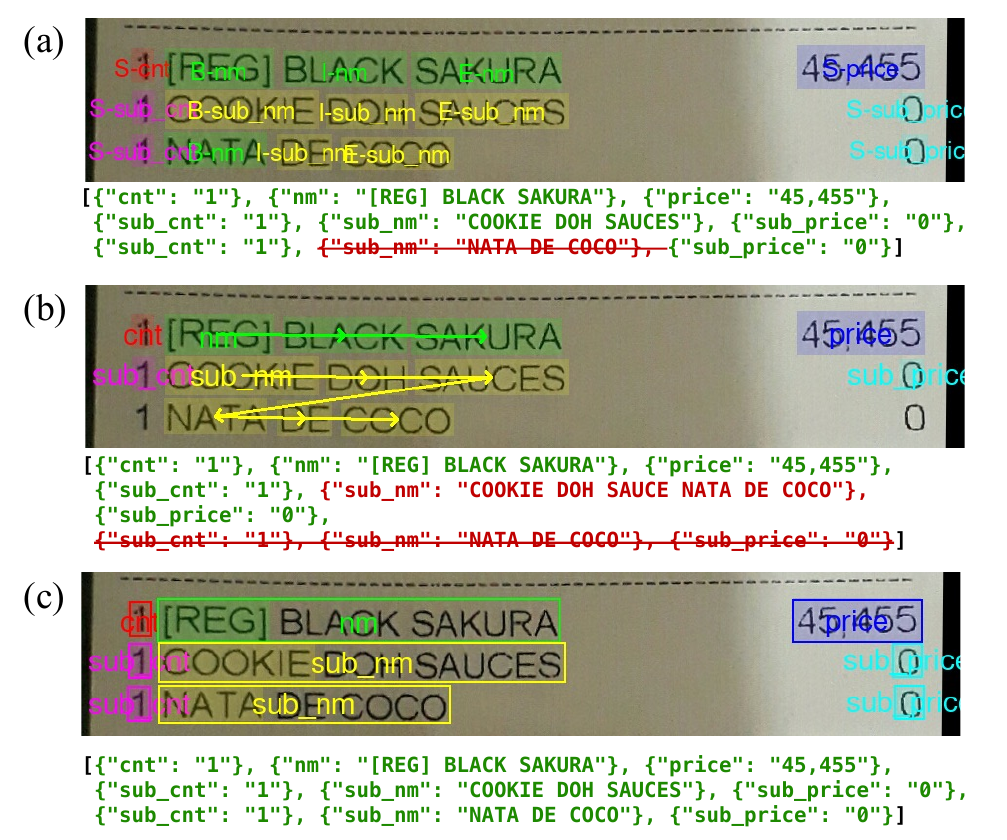}
  \vspace{-1.5em}
  \captionof{figure}{Visualization of receipt parsing results using different SIE formulations. Each result consists of the visualization of model predictions, and the parsing outputs (given the model predictions). \textbf{(a) IOB tagging} visualizes the predicted tags of OCR words. \textbf{(b) SPADE decoding} visualizes the decoded graph, and arrows between words indicate that the words are linked in the same entity. \textbf{(c) DocTr} visualizes the predicted anchor words and their bounding boxes. For the parsing outputs, green/red text means the predicted text matches/does not match ground truth. Strikethrough text means the ground truth text is missed from prediction.}
  \label{fig:cord_parsing_comparison} \vspace{-1em}
\end{figure}

\subsection{Model Properties} \label{sec:alation_study}
We analyze DocTr's design and consider other choices.
\vspace{-1em}
\paragraph{Problem Formulation.} We use DocTr as the backbone network for the encoding of document inputs (image and OCR words) and apply different formulations to decode structured information. Specifically, we compare our formulation with IOB tagging and graph based solutions. For IOB tagging, we follow the literature~\cite{lee2022formnet,xu2020layoutlm} and assign BIOES tags to each token and decode entities according to the tagged entity spans. Note that IOB tagging does not support entity linking. For a fair comparison, we link entities using a way similar to the anchor word association method introduced in Sec.~\ref{sec:doctr}. We treat the ``B'' tag or ``S'' tag of entities as the anchor words and link entities via decoding of entity linking affinity matrices. For graph based SIE, we follow the literature~\cite{hong2021bros,hwang2020spatial} by attaching a SPADE~\cite{hwang2020spatial} decoder at the end of DocTr. We fine-tune DocTr and decode graphs using the same way as specified in the original SPADE method. To understand the sensitivity of the SIE formulations with regard to the reading orders of input text, we evaluate them under two text serialization settings, raster scan and oracle. For oracle, we first order the ground truth entities in a raster scan manner, then order text while preserving the entity order.

Table~\ref{tab:formulation} shows the receipt parsing results on the CORD dataset. Our proposed formulation achieves the best performance in both text serialization settings. We notice that, compared with the other two formulations, our formulation is less sensitive to text serialization with only 0.6 score drop (vs. 0.9 drop by IOB tagging or SPADE) while switching from oracle to raster scan text serialization. We also observe that our formulation can better address cases where there is dense text with multiple entities near each other. Fig.~\ref{fig:cord_parsing_comparison} shows an example visualization (see supplementary material for more results). For IOB tagging, it can tag most of the words well. However, even a single tagging error can cause failures of entity decoding, and an entity is missed from the parsing outputs. For SPADE, the dense words result in a challenge for constructing an entity graph, and the model incorrectly merges the two \texttt{sub\_nm}'s as a single entity. In comparison, DocTr only requires identifying the anchor words which is an easier task and, with bounding box predictions, all the entities are correctly extracted.

\begin{table}[!t]
    \centering
    \tablestyle{4pt}{1.05}
    \begin{tabular}{cccc}
    \bf anchor word  &  \bf primary anchor & \bf parsing (C) \\
    \midrule
    first            &   name       &    94.2     \\
    last             &   name       &    94.1     \\
    first + last     &   first      &    94.0     \\
    first + last     &   name      &  \bf{94.4}   \\
    \end{tabular}
    \vspace{-0.7em}
    \caption{Receipt parsing (CORD) results under different choices of anchor words and primary anchors.
    }
    \label{tab:anchor_word} \vspace{-1em}
\end{table}

\vspace{-1em}
\paragraph{Anchor Word and Primary Anchor.} We investigate different ways of designating anchor word and primary anchor. In Sec.~\ref{sec:new_formulation}, we introduced using the first word (in terms of reading order) of an entity as the anchor word. Here, we consider two alternatives: 1) using the last word or 2) both the first and last word as the anchor. Table~\ref{tab:anchor_word} (row 1-2, 4) shows the comparison of these three choices. We notice that there is no significant differences (94.2 vs. 94.1) between using the first word and last word as the anchor word. Using both first and last as the anchor word gives slightly better performance. We hypothesize that this is because first and last words help better identify the boundary of an entity.

For primary anchor, we investigate its choices for line-item extraction. We consider two candidates: 1) using the anchor word of the first entity in a line-item, or 2) using the anchor word of \texttt{name} as the primary anchor. From Table~\ref{tab:anchor_word} (row 3 and 4), we see the latter is a better choice with 0.4 improvement. This is reasonable since the first entity in a line-item may vary semantically (i.e., it could be \texttt{name}, \texttt{cnt} or other entity types), and thus it is harder to identify. However, this choice is also more flexible than using \texttt{name} as the primary anchor because there may be no \texttt{name} in an line-item. For CORD, each line-item always has a \texttt{name}, so this is not a concern (see supplementary material for primary anchor choices of other entity categories).

\begin{table}[!t]
    \centering
    \tablestyle{4pt}{1.05}
    \begin{tabular}{ccccc}
    \bf pre-training            & \bf parsing (C) & \bf ELB (F) & \bf ELK (F) \\
    \midrule
     none                       & 82.3            &  14.2      & 12.0       \\
     MVLM \cite{xu2020layoutlm} & 90.9            &  82.7      & 73.0       \\
     MDM                        & \bf{94.4}       &  \bf 84.0  & \bf 73.9  \\
    \end{tabular}
    \vspace{-0.7em}
    \caption{Receipt parsing (CORD), entity labeling (FUNSD) and entity linking (FUNSD) results using different pre-training tasks.}
    \label{tab:pretraining_tasks} \vspace{-1em}
\end{table}

\begin{figure}[t]
  \centering
  \includegraphics[width=\linewidth]{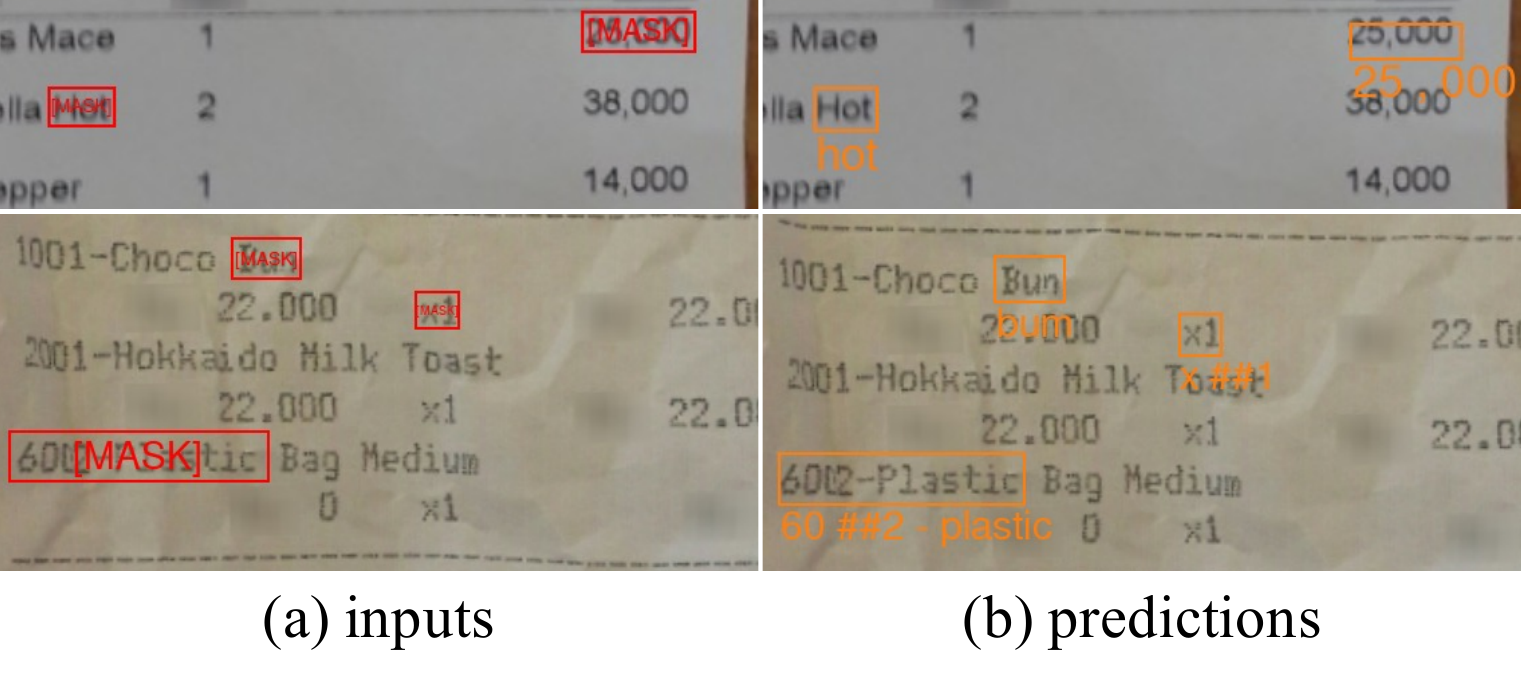}
  \vspace{-2.1em}
  \caption{Example pre-training predictions on CORD images. For inputs, we visualize masked word boxes, and their text is replace by \texttt{[MASK]}. For predictions, we visualize the predicted word boxes of the masked inputs. Under each box prediction, we also visualize its corresponding word token predictions.}
  \label{fig:pretraining_outputs} \vspace{-1em}
\end{figure}

\vspace{-1em}
\paragraph{Pre-training.} We evaluate the effectiveness of the pre-training task (MDM) introduced in Sec.~\ref{sec:training}. We consider three settings: 1) without pre-training, 2) with MVLM and 3) with MDM. Table~\ref{tab:pretraining_tasks} compares their performances. Without pre-training, the performance drops significantly. With MVLM, the performance improves but still falls behind using MDM. This shows the effectiveness of having MDM for document understanding pre-training. In particular, we see more benefit of using MDM for receipt parsing. This is because our proposed formulation requires bounding box regression, and \textit{MDM helps learn better box predictions}.

We also show example MDM pre-training predictions in Figure~\ref{fig:pretraining_outputs}. Note that since both the input OCR box and text are masked, the model will need to not only predict what is masked but also predict where to find the masked word. We can see in most of the cases, the model can predict both kinds of information well. There are cases where the box (e.g., ``25,000'' in row 1) or text (e.g., ``Bun'' in row 2) is not accurately predicted. But the errors are reasonable. We also notice that the model can predict words that cannot be inferred through only text context, such as prices. This shows \textit{the usage of visual information}.

\begin{table}[!t]
    \centering
    \tablestyle{4pt}{1.05}
    \begin{tabular}{cccccccc}
    & \bf vis. enc. & \bf VL-dec.     & \bf LCQ    & \bf parsing (C) & \bf ELB (F) & \bf ELK (F) \\
    \midrule
    1) &            &                 & N/A        & 92.3            &  82.1      & 73.2       \\
    2) &            & \checkmark      & \checkmark & 92.6            &  83.0      & 73.3       \\
    3) & \checkmark & \checkmark      &            & 90.7            &  14.9      & 9.5        \\
    4) & \checkmark & \checkmark      & \checkmark & \bf{94.4}       &  \bf 84.0 & \bf 73.9       \\
    \end{tabular}
    \vspace{-0.7em}
    \caption{Impact of different architectural components: vision encoder, VL-decoder and language conditioned queries. \checkmark means the component is included in the architecture.}
    \label{tab:architecture_ablation} \vspace{-1em}
\end{table}

\vspace{-1em}
\paragraph{Architecture Design.} We then ablate the impact of the architectural components of DocTr. Table~\ref{tab:architecture_ablation} shows the ablation results. The first row is a DocTr model with only the language encoder which is equivalent to the LayoutLM~\cite{xu2020layoutlm} model without visual inputs. The second row is a model with both the language encoder and VL-decoder but no vision encoder. These two models are close in performance. This is reasonable as without visual inputs the VL-decoder does not add much of information for decoding. Row 4 is the full model with both the vision encoder and VL-decoder. Compared with row 1 and 2, the performance improves noticeably. This suggests the importance of using visual information.

For row 3 and 4, we study the effectiveness of using the proposed language conditioned queries (LCQ). Specifically, we apply Eq.~(\ref{eq:language_conditioined_ca}) to the cross-attention module when LCQ is checked. Otherwise, the standard cross-attention is used. We can see that LCQ is important since it helps to guide this one-to-one mapping between OCR and outputs, which is required by our proposed formulation. 
\section{Conclusion}

We have presented a new approach for SIE from visually-rich documents. This approach is based on our novel formulation which includes object detection as part of the problem setting. This naturally leads us to include a transformer-based object detector as part of the architecture design and an object detection based loss in pre-training. 

We have empirically shown that our proposed object detection based formulation readily addresses the structured information extraction task, and our solution outperforms existing solutions on SIE benchmarks. We hope this approach will initiate more efforts in combining object detection with existing vision-language models for document intelligence.

We also note that using anchor words limits the application of this approach to text-rich documents, and text-based entity extraction only. For future work, we explore solutions that extend the propose formulation for the extraction of non-textual content (e.g., symbols, logos, etc.) from documents.

{\small
\bibliographystyle{ieee_fullname}
\bibliography{references}
}

\clearpage

\appendix
\section{Appendix}
\subsection{Implementation Details}
\graphicspath{ {./images/supp/} }

The hyperparameters we used for pre-training and fine-tuning downstream tasks are shown in Table~\ref{tab:settings}. In general, our hyperparameter settings are similar to the one
used in Deformable DETR~\cite{zhu2020deformable}. Here, ``base'' configurations are those common for all experiments. It is worth mentioning that we use the box refinement design proposed in \cite{zhu2020deformable} which we find helpful for bounding box prediction.

For ``pre-training'', we use a three-stage pre-training. That is, in the first stage, we pre-train on 1M IIT-CDIP samples~\cite{lewis2006building} for 20 epochs. Then, we pre-train on 5M samples for 5 epochs. In the final stage, we pre-train on 11M samples (full dataset) for 2 epochs.

``receipt parsing'', ``entity labeling'' and ``entity linking'' show the settings we used to obtain the numbers we reported in Table 1 and 2 of the main manuscript. For ``entity extraction'', since we follow the existing work to address this task with IOB tagging (for fair comparison), we do not apply Eq. (2) and Eq. (3) (in main manuscript) as the loss function. But instead we simply use a cross-entry loss. Here ``CE loss weight'' is the weight of this loss.

\begin{table}[h]
    \centering
    \tablestyle{4pt}{1.05}
    \begin{tabular}{ccc}
    \bf config type & \bf config name & \bf value \\
    \midrule
    \multirow{7}{*}{\bf base}   &    optimizer              & AdamW~\cite{loshchilov2018decoupled}          \\ 
                                &    base LR                & $2e^{-4}$ \\
                                &    cnn  LR                & $2e^{-5}$ \\
                                &    language encoder LR.   & $1e^{-5}$ \\
                                &    weight decay           & $1e^{-4}$ \\
                                &    LR schedule            & step      \\
                                &    box refinement~\cite{zhu2020deformable} & yes       \\
    \midrule
    \multirow{4}{*}{\bf pre-training} &    batch size             & 32          \\
                                &    epochs                 & 20, 5, 2    \\
                                &    training samples       & 1M, 5M, 11M \\
                                &    LR drop step size      & 20          \\
    \midrule                
    \multirow{7}{*}{\bf receipt parsing} &    batch size             & 8           \\
                                &    epochs                 & 200         \\
                                &    LR drop step size      & 160         \\
                                &    EE loss weight         & 5.0         \\
                                &    EL loss weight         & 1.0         \\
                                &    anchor word            & first+last  \\
                                &    primary anchor         & name        \\
    \midrule
    \multirow{4}{*}{\bf entity labeling} &    batch size             & 8           \\
                                &    epochs                 & 50          \\
                                &    LR drop step size      & 40          \\
                                &    CE loss weight         & 2.0         \\
    \midrule
    \multirow{5}{*}{\bf entity linking} &    batch size             & 8           \\
                                &    epochs                 & 200         \\
                                &    LR drop step size      & 160         \\
                                &    EE loss weight         & 0.0         \\
                                &    EL loss weight         & 10.0        \\
                    
    \end{tabular}
    \vspace{-0.7em}
    \caption{Implementation settings. Here, ``LR'' stands for learning rate. ``EE'' stands for entity extraction. ``EL'' stands for entity linking. ``CE'' stands for cross-entropy.}
    \label{tab:settings}
\end{table}

\subsection{VL Decoder Layer}

\begin{figure}[t]
  \centering
  \includegraphics[width=0.85\linewidth]{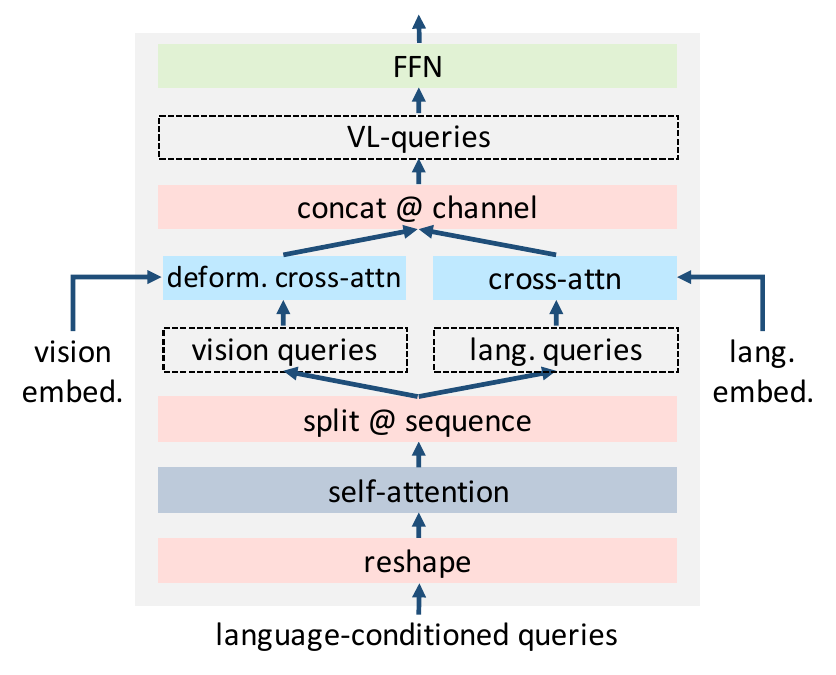}
  \captionof{figure}{Architecture of a VL-decoder layer. It splits language-conditioned queries into two parts, vision queries and language queries. Vision queries extract visual information via deformable cross-attention~\cite{zhu2020deformable}. Language queries extract linguistic information via language-conditioned cross-attention. }
  \label{fig:decoder}
\end{figure}

The VL-decoder has 6 layers, where each layer consists of a self-attention module, a deformable cross-attention module~\cite{zhu2020deformable} and a standard cross-attention module~\cite{vaswani2017attention} as shown Fig.~\ref{fig:decoder}. Let $\mathbf{Q} \in \mathbb{R}^{L \times D}$ be the input decoder queries. We first reshape $\mathbf{Q}$ to $\mathbf{Q'} \in \mathbb{R}^{2L \times \frac{D}{2}}$, where $L$ is the number of input queries and $D$ is the channel size. After the self-attention module, we split $\mathbf{Q'}$ into two equally sized queries $\mathbf{Q}^{v} \in \mathbb{R}^{L \times \frac{D}{2}}$ and $\mathbf{Q}^{l} \in \mathbb{R}^{L \times \frac{D}{2}}$. The vision queries $\mathbf{Q}^{v}$ extract visual features via deformable cross-attention. The language queries $\mathbf{Q}^{l}$ extract language features via cross-attention, where we apply Eq.~(1) in the main text to assign explicit language semantics to queries. The outputs from the two cross-attention modules are concatenated at the channel dimension to recover the original shape, i.e., $\mathbf{Q}^{vl} \in \mathbb{R}^{L \times D} = \text{concat}(\mathbf{Q}^{v}, \mathbf{Q}^{l})$. We use a fully connected layer at the end to further fuse vision and language information along the channel dimension.

\subsection{Additional Results}

\begin{figure*}[t]
  \centering
  \includegraphics[width=0.8\linewidth]{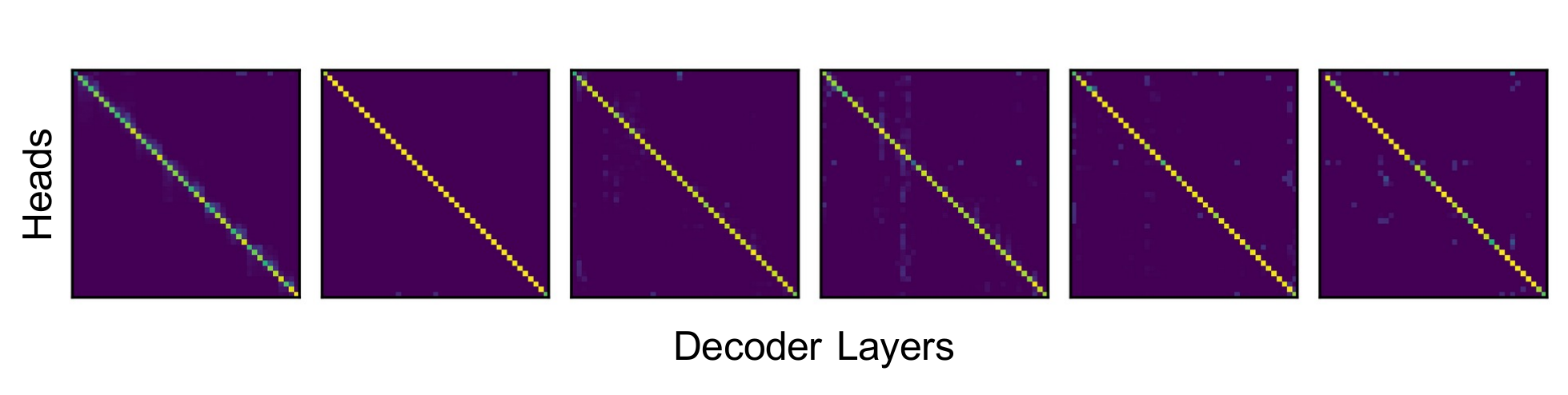}
  \caption{Visualization of language-conditioned cross-attention. }
  \label{fig:pretraining}
\end{figure*}

\paragraph{Comparison of using anchor words from different
line-item fields as primary anchors.} From Table~\ref{tab:primary_anchor_comparison}, we can see that using the anchor word of ``name'' gives the best result. This is because in the test set, all the line-items contain this field. So, it is reliable to use this field as the primary anchor for entity linking. For the other fields, their performances are lower when they do not frequently present in line-items. Also, note that the middle column only considers line-items not key-values. Therefore, it is possible that the parsing performance numbers are even higher than the proportion of line-items that contain this field.

\begin{table}[!t]
    \centering
    \tablestyle{4pt}{1.05}
    \begin{tabular}{ccc}
    \bf primary anchor & \bf LI's with this field & \bf parsing (CORD) \\
    \midrule
     unit price      & 28\%    &  66.5        \\ 
     count           & 90\%    &  91.4        \\ 
     price           & 99\%    &  94.0        \\ 
     name            & 100\%   &  \bf{94.4} \\ 
    \midrule
    first            & -       &  94.0      \\
    \end{tabular}
    \vspace{-0.7em}
    \caption{Receipt parsing results using anchor words from different line-item fields as primary anchors. The middle column indicates the proportion of line-items that contain this field.}
    \label{tab:primary_anchor_comparison}
\end{table}

\paragraph{Comparison of using predicted text and ground truth text as inputs.} The existing works~\cite{hwang2020spatial,hong2021bros,xu2020layoutlm,cui2021document} use ground truth text as the input in the experiments. We also follow the same way for fair comparison. However, it would be interesting to see how the models work if predicted text (from an OCR system) is used. In particular, using ground truth text as input is not in favor of vision only approaches such as Donut~\cite{kim2022ocr}. Table~\ref{tab:pred_vs_gt_ocr} shows the results of the comparison. In this experiment, we use an in-house OCR system which has comparable performances with the state-of-the-art OCR solutions (e.g., those from Azure, GCP or AWS). As we can see, there is a performance drop when we switch from using ground truth text to predicted text. However, compared with the vision only solution Donut, we are still noticeably better. This indicates the importance of having language inputs.

\paragraph{Model performance at different pre-training stages.} In Table~\ref{tab:pretrain}, each stage is based on the pre-trained model from its previous stage. For example, stage 2 initializes the model using the weights pre-trained from stage 1. As we can see, when more data is used, the model's performance continues improving on the FUNSD entity extraction task.

\paragraph{Visualization of language-conditioned cross-attention.} We further verify the behavior of this cross-attention mechanism by visualizing the cross-attention matrices. The cross-attention results are extracted from the cross-attention module of each decoder layer, i.e., we check the cross-attention between the language inputs and the language-conditioned queries. As we can see in Fig~\ref{fig:pretraining}, the attention weights are high on the diagonal of the attention matrices. This shows that we successfully established this one-to-one mapping between the queries (and thus decoder outputs) and language tokens.

\begin{figure*}
  \centering
  \includegraphics[width=\linewidth]{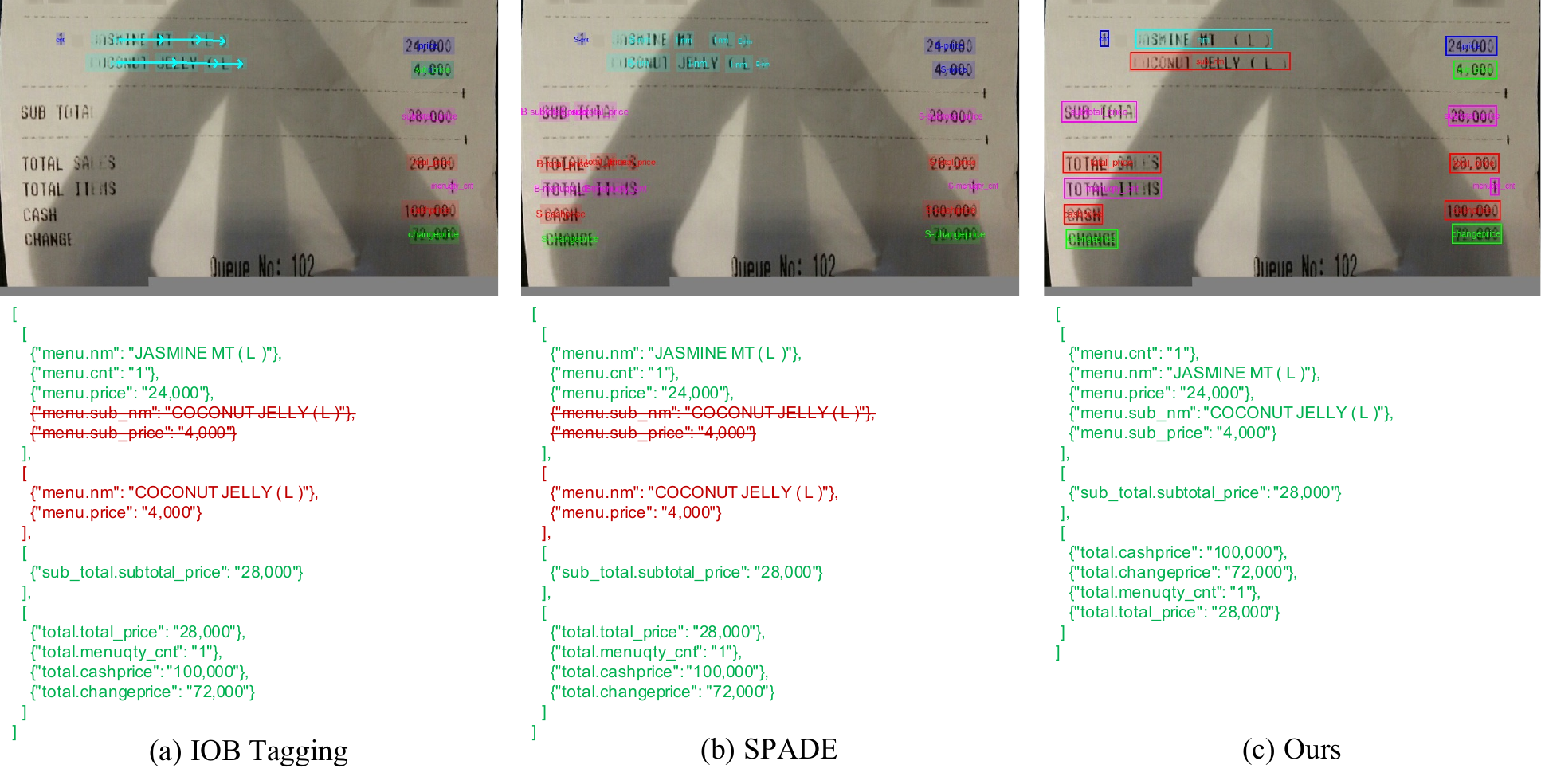}
  \caption{CORD receipt parsing results on \textbf{receipt\_00002} sample. Each result consists of the visualization of model predictions, and the parsing outputs. \textbf{(a)} IOB tagging visualizes the predicted tags of OCR words. \textbf{(b)} SPADE visualizes the decoded graph, and arrows between words indicate that the words are linked in the same entity. \textbf{(c)} Ours visualizes the predicted anchor words and their bounding boxes. For simplicity of the visualization,  entity-linking results are not visualized in here. For the parsing outputs, \textcolor{green}{green}/\textcolor{red}{red} text means the predicted text matches/does not match ground truth. \textcolor{red}{\st{Strikethrough}} text means the ground truth text is missed from prediction. Best view in color and zoom-in for details of the visualization. For this example, both IOB tagging and SPADE recognized the second row of the line-item as an individual line-item. DocTr understands the line-item better by recognizing it as a single line-item.}
  \label{fig:parsing_1}
\end{figure*}

\begin{figure*}
  \centering
  \includegraphics[width=\linewidth]{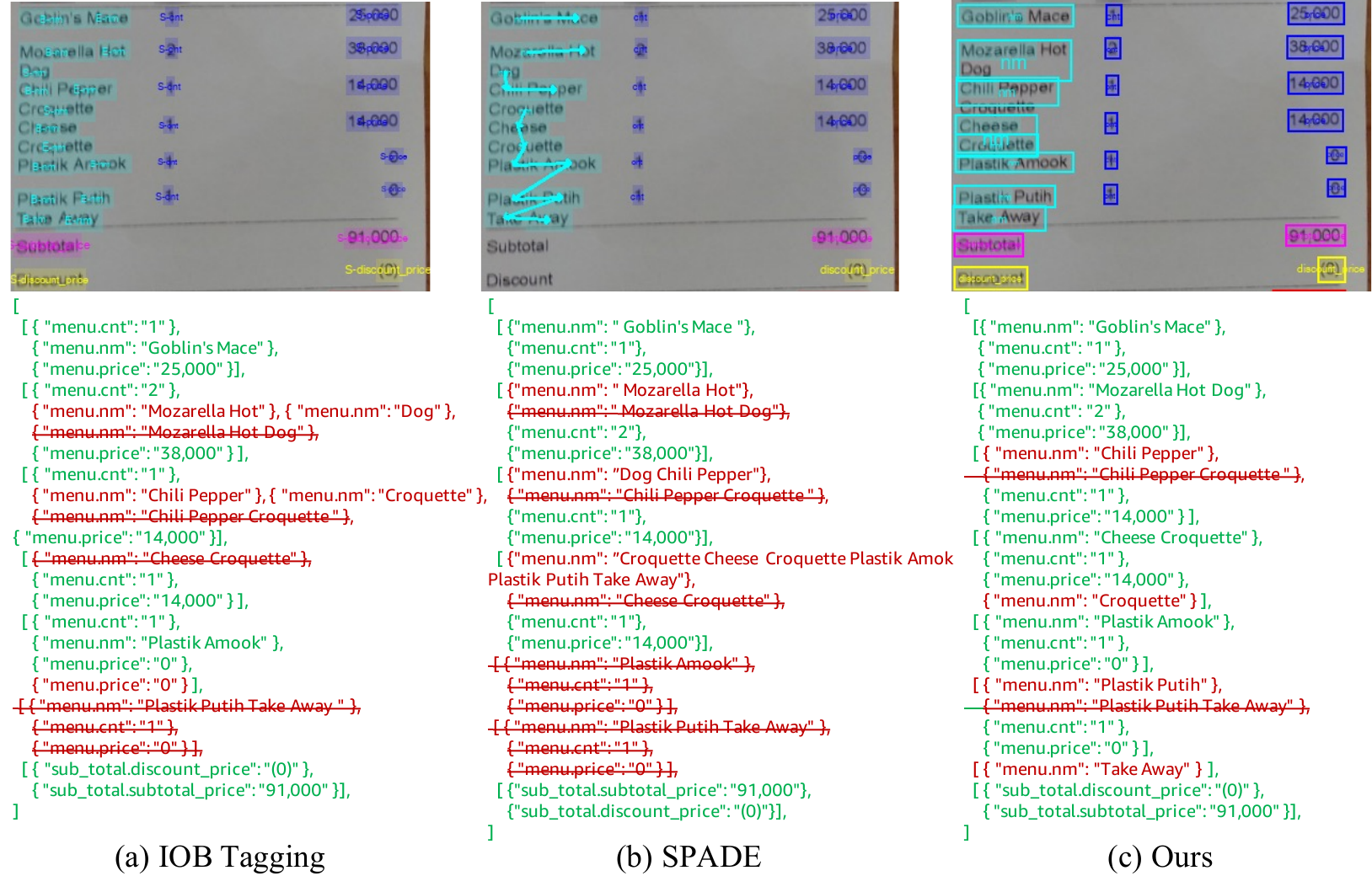}
  \caption{CORD receipt parsing results on \textbf{receipt\_00017} sample. Each result consists of the visualization of model predictions, and the parsing outputs. \textbf{(a)} IOB tagging visualizes the predicted tags of OCR words. \textbf{(b)} SPADE visualizes the decoded graph, and arrows between words indicate that the words are linked in the same entity. \textbf{(c)} Ours visualizes the predicted anchor words and their bounding boxes. For simplicity of the visualization,  entity-linking results are not visualized in here. For the parsing outputs, \textcolor{green}{green}/\textcolor{red}{red} text means the predicted text matches/does not match ground truth. \textcolor{red}{\st{Strikethrough}} text means the ground truth text is missed from prediction. Best view in color and zoom-in for details of the visualization. This is a challenging sample with several line-items. Some line-items have single-line name, and some have two-line name. As we can see, SPADE totally failed in this case, with a wrong graph for the line-item names. IOB tagging and DocTr are better and detected names mostly correct.}
  \label{fig:parsing_2}
\end{figure*}

\begin{figure*}
  \centering
  \includegraphics[width=\linewidth]{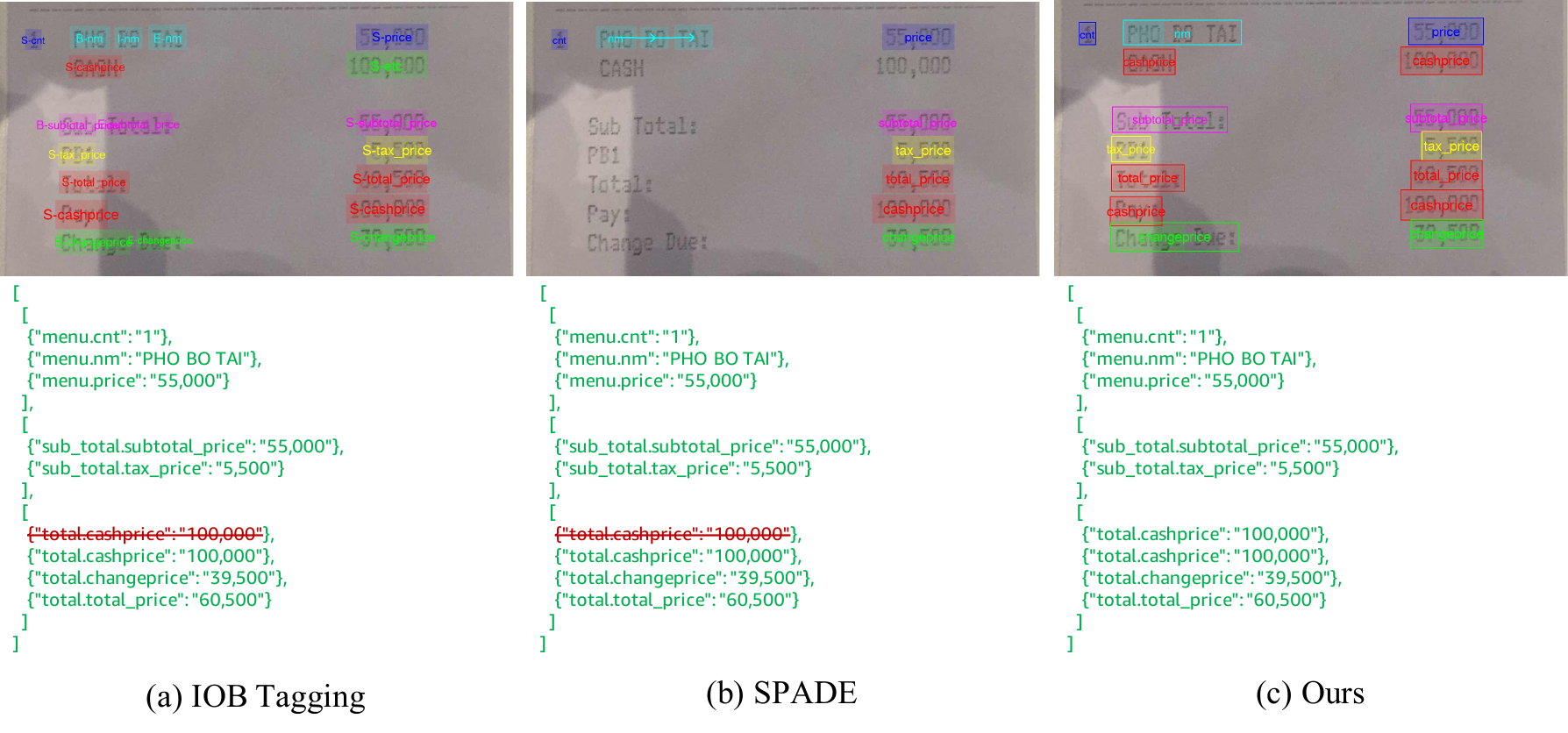}
  \caption{CORD receipt parsing results on \textbf{receipt\_00058} sample. Each result consists of the visualization of model predictions, and the parsing outputs. \textbf{(a)} IOB tagging visualizes the predicted tags of OCR words. \textbf{(b)} SPADE visualizes the decoded graph, and arrows between words indicate that the words are linked in the same entity. \textbf{(c)} Ours visualizes the predicted anchor words and their bounding boxes. For simplicity of the visualization,  entity-linking results are not visualized in here. For the parsing outputs, \textcolor{green}{green}/\textcolor{red}{red} text means the predicted text matches/does not match ground truth. \textcolor{red}{\st{Strikethrough}} text means the ground truth text is missed from prediction. Best view in color and zoom-in for details of the visualization. In this example, ``CASH'' is a subtotal. But it is right below the line-item. Thus, both ``IOB Tagging'' and ``SPADE'' missed this detection.}
  \label{fig:parsing_3}
\end{figure*}

\begin{figure*}
  \centering
  \includegraphics[width=\linewidth]{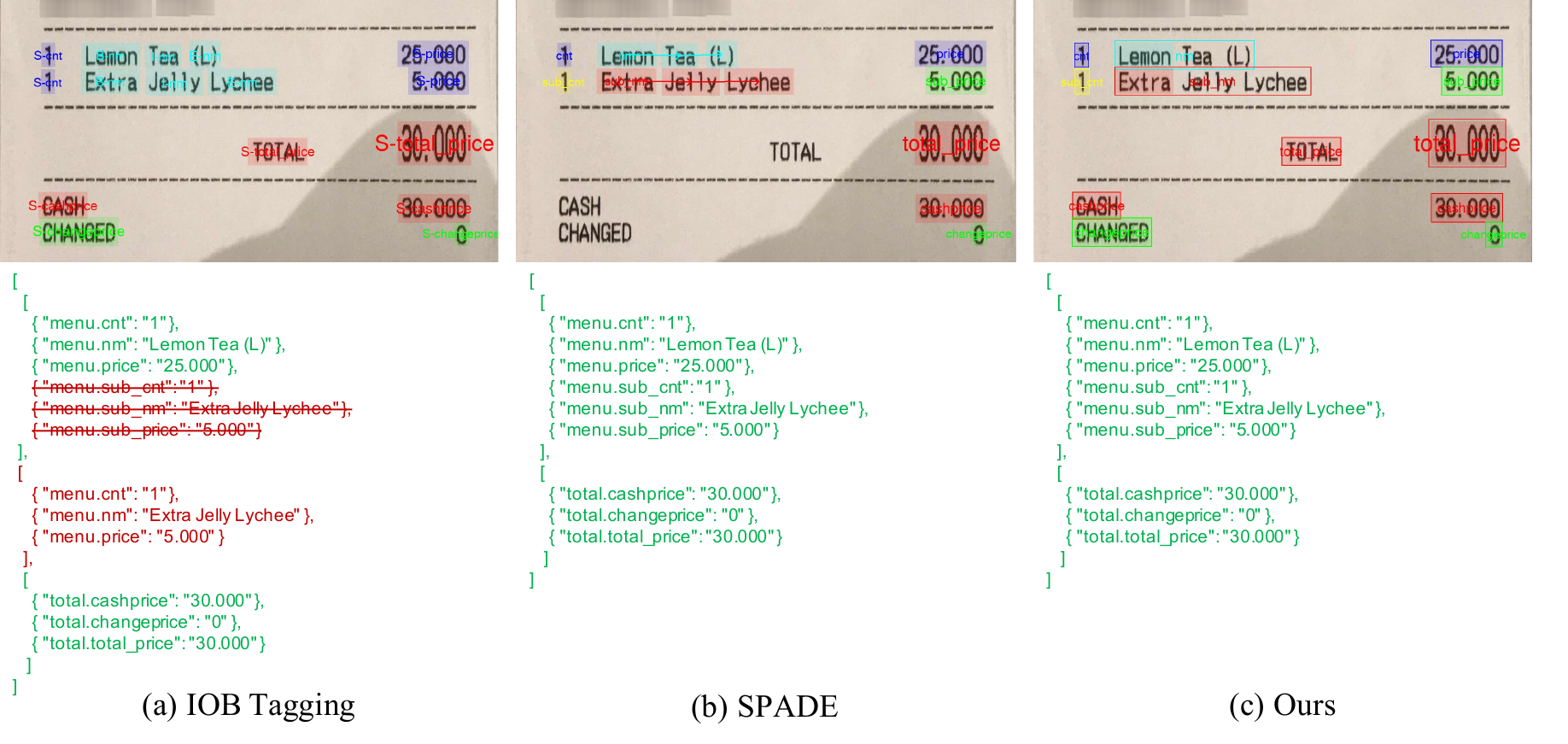}
  \caption{CORD receipt parsing results on \textbf{receipt\_00078} sample. Each result consists of the visualization of model predictions, and the parsing outputs. \textbf{(a)} IOB tagging visualizes the predicted tags of OCR words. \textbf{(b)} SPADE visualizes the decoded graph, and arrows between words indicate that the words are linked in the same entity. \textbf{(c)} Ours visualizes the predicted anchor words and their bounding boxes. For simplicity of the visualization,  entity-linking results are not visualized in here. For the parsing outputs, \textcolor{green}{green}/\textcolor{red}{red} text means the predicted text matches/does not match ground truth. \textcolor{red}{\st{Strikethrough}} text means the ground truth text is missed from prediction. Best view in color and zoom-in for details of the visualization.}
  \label{fig:parsing_4}
\end{figure*}

\begin{figure*}
  \centering
  \includegraphics[width=0.9\linewidth]{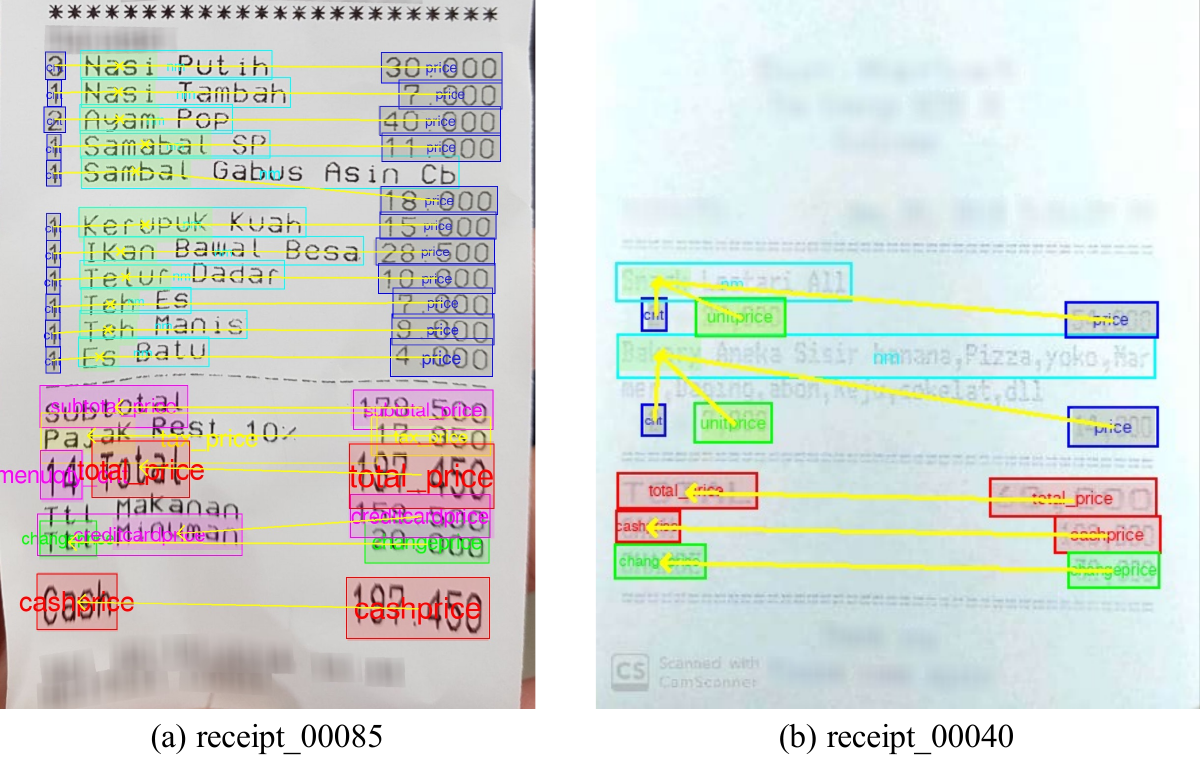}
  \caption{CORD receipt parsing failure cases. We visualize the predicted anchor words, their bounding boxes, and anchor word associations (yellow arrows).}
  \label{fig:failure_cases}
\end{figure*}

\begin{figure*}
  \centering
  \includegraphics[width=\linewidth]{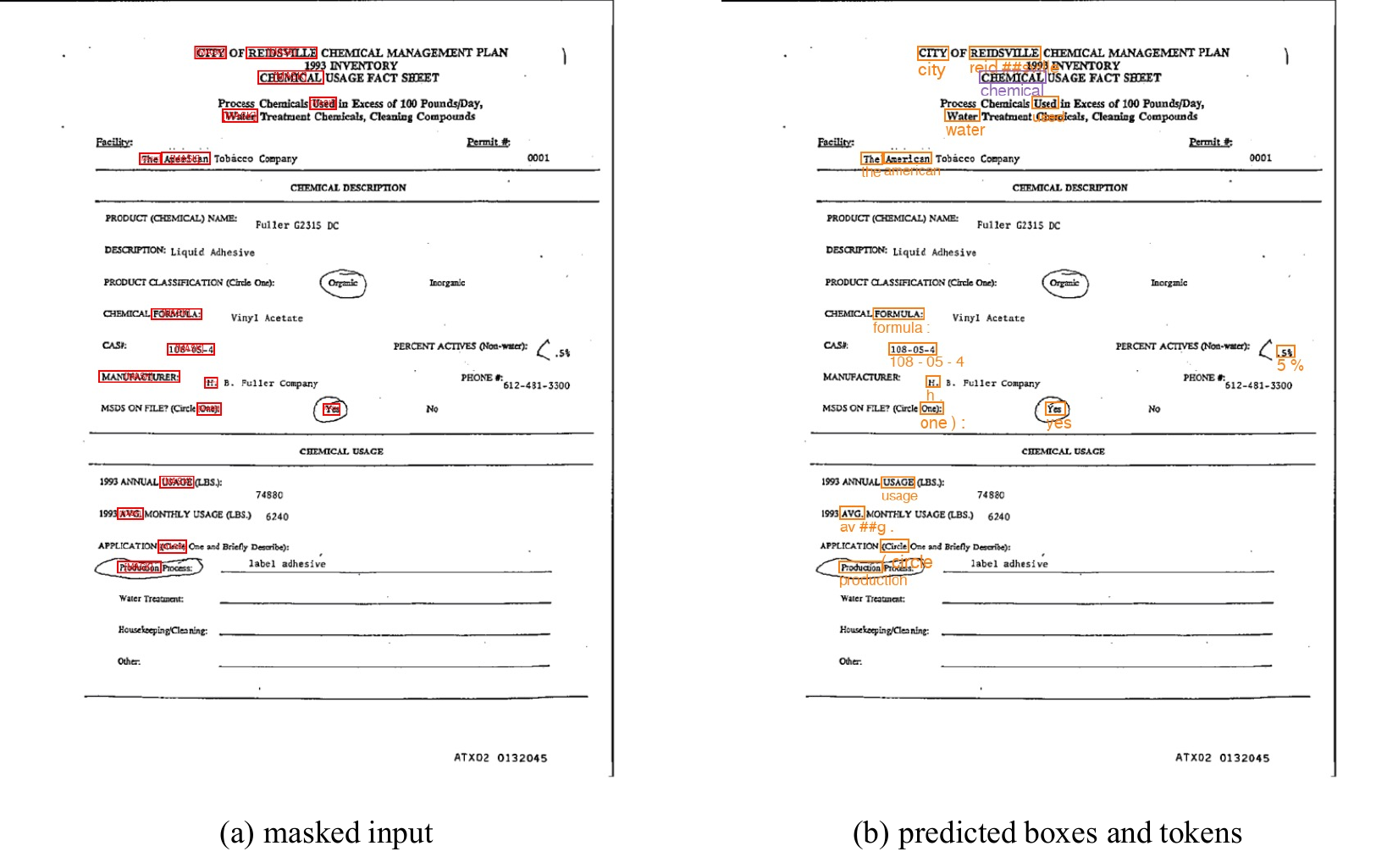}
  \caption{Example pre-training predictions on FUNSD sample \textbf{0060024314}. For inputs, we visualize masked word boxes, and their text is replace by \texttt{[MASK]}. For predictions, we visualize the predicted word boxes of the masked inputs. Under each box prediction, we also visualize its corresponding word token predictions.}
  \label{fig:pretrain_1}
\end{figure*}

\begin{figure*}
  \centering
  \includegraphics[width=\linewidth]{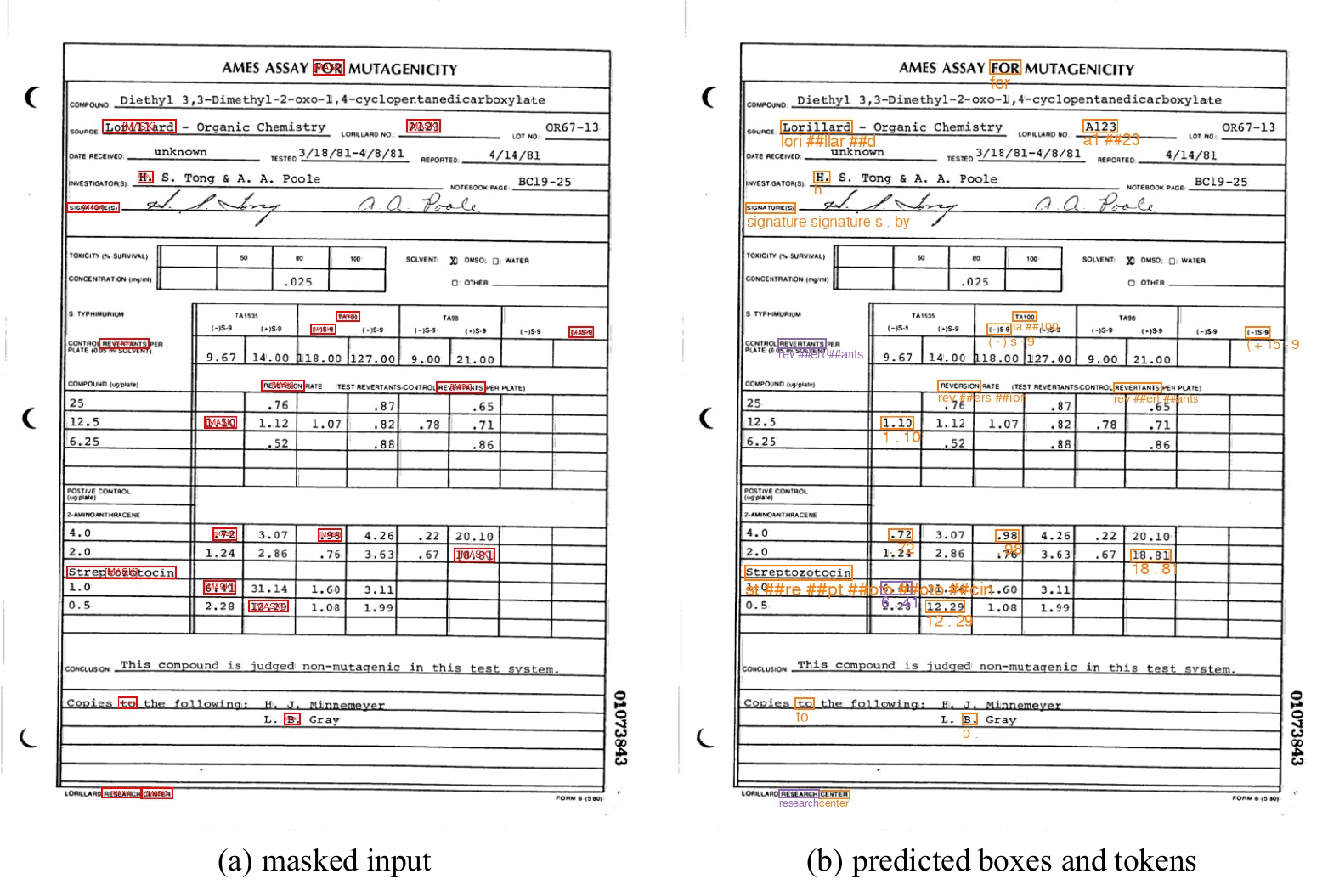}
  \caption{Example pre-training predictions on FUNSD sample \textbf{01073843}. For inputs, we visualize masked word boxes, and their text is replace by \texttt{[MASK]}. For predictions, we visualize the predicted word boxes of the masked inputs. Under each box prediction, we also visualize its corresponding word token predictions.}
  \label{fig:pretrain_2}
\end{figure*}

\begin{figure*}
  \centering
  \includegraphics[width=\linewidth]{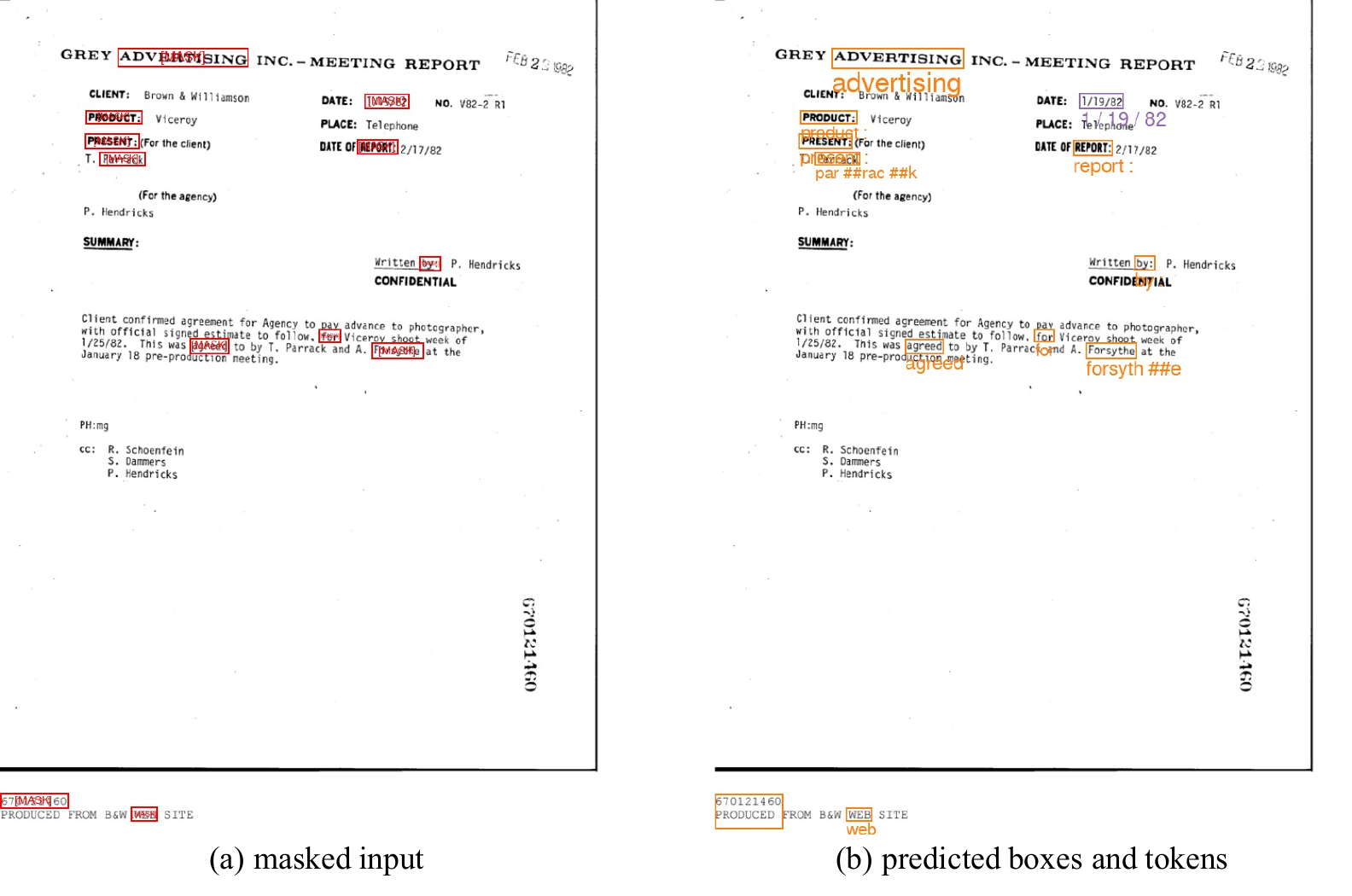}
  \caption{Example pre-training predictions on FUNSD sample \textbf{660978}. For inputs, we visualize masked word boxes, and their text is replace by \texttt{[MASK]}. For predictions, we visualize the predicted word boxes of the masked inputs. Under each box prediction, we also visualize its corresponding word token predictions.}
  \label{fig:pretrain_3}
\end{figure*}

\begin{figure*}
  \centering
  \includegraphics[width=\linewidth]{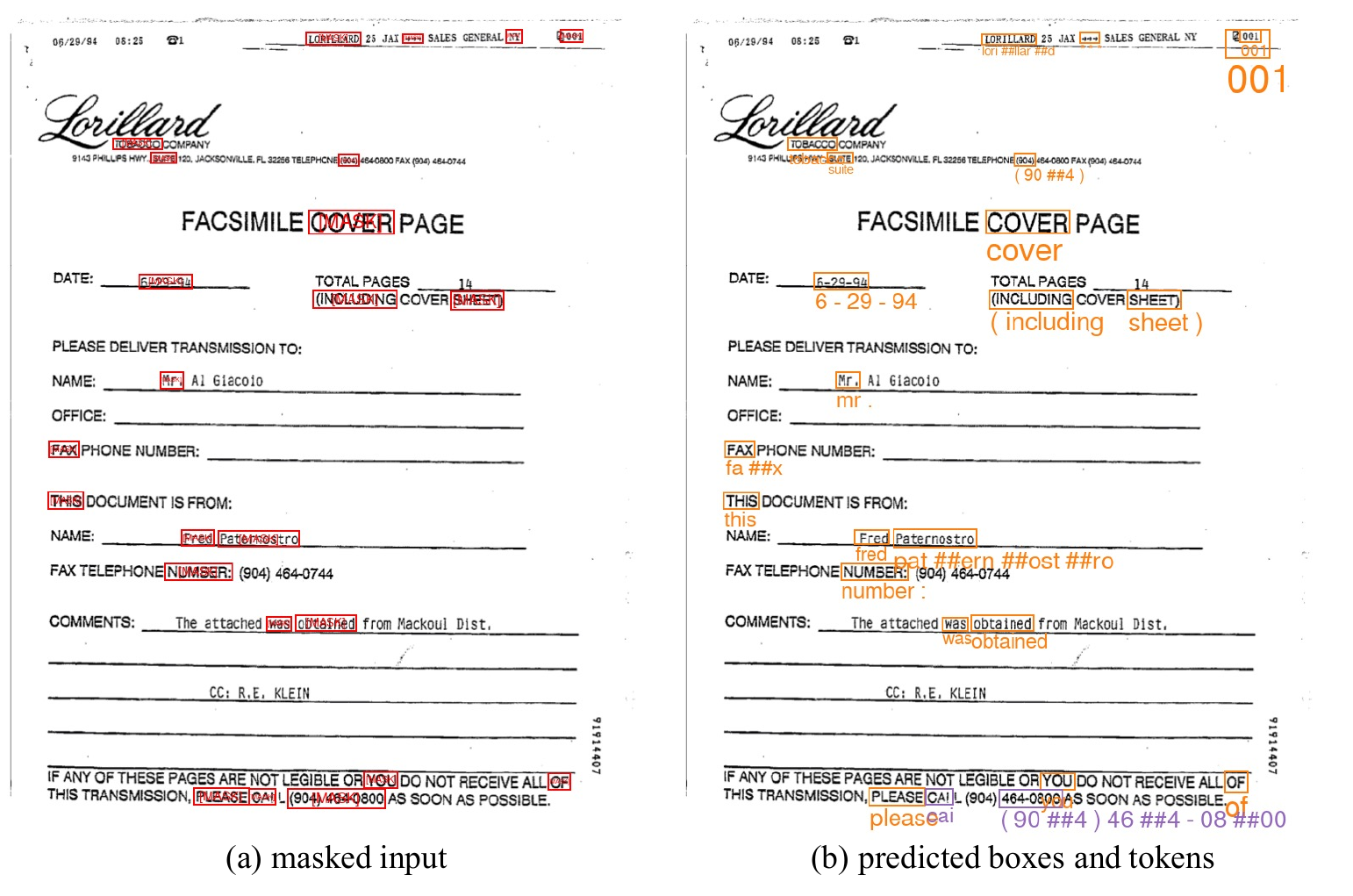}
  \caption{Example pre-training predictions on FUNSD sample \textbf{91914407}. For inputs, we visualize masked word boxes, and their text is replace by \texttt{[MASK]}. For predictions, we visualize the predicted word boxes of the masked inputs. Under each box prediction, we also visualize its corresponding word token predictions.}
  \label{fig:pretrain_4}
\end{figure*}

\paragraph{Additional visualizations.} Fig~\ref{fig:parsing_1}-\ref{fig:parsing_4} shows additional comparisons of three structured information extraction formulations. Fig~\ref{fig:failure_cases} shows two receipts parsing failure cases. Fig~\ref{fig:pretrain_1}-\ref{fig:pretrain_4} shows additional pre-training outputs using our proposed masked detection modeling task.

\begin{table}[!t]
    \centering
    \tablestyle{4pt}{1.05}
    \begin{tabular}{cccc}
    \bf model                                        & \bf input text & \bf parsing (CORD) \\
    \midrule
    $\textrm{Donut}$~\cite{kim2022ocr}               & none           & 87.8 \\
    $\textrm{SPADE}$~\cite{hwang2020spatial}         & gt             & 92.5 \\
    \midrule
    $\textrm{LayoutLMv2}$~\cite{xu2021layoutlmv2}    & pred.  & 92.2 \\
    $\textrm{BROS}$~\cite{hong2021bros}              & pred.  & 92.1 \\
    $\textrm{LayoutLMv3}$~\cite{huang2022layoutlmv3} & pred.  & 93.0 \\
    $\textrm{DocTr (ours)}$                          & pred.  & 93.7 \\
    \midrule
    $\textrm{LayoutLMv2}$~\cite{xu2021layoutlmv2}    & gt  & 92.7 \\
    $\textrm{BROS}$~\cite{hong2021bros}              & gt  & 92.9 \\
    $\textrm{LayoutLMv3}$~\cite{huang2022layoutlmv3} & gt  & 93.6 \\
    $\textrm{DocTr (ours)}$                          & gt  & \bf{94.4} \\
    \end{tabular}
    \vspace{-0.7em}
    \caption{Receipt parsing results comparison of using predicted text (from OCR system) and ground truth text as the input to the model. ``gt'' means ground truth text is used (but the text is ordered in raster scan manner). ``pred.'' means predicted text is used. ``none'' means no text input is used.}
    \label{tab:pred_vs_gt_ocr}
\end{table}

\begin{table}[!t]
    \centering
    \tablestyle{4pt}{1.05}
    \begin{tabular}{cccc}
    \bf stage   & \bf \# samples & \bf \# epochs                          & \bf EE (FUNSD) \\
    \midrule
        1       & 1M                         & 20 epochs                  & 82.1            \\
        2       & 5M                         & 5 epochs                   & 83.1            \\
        3       & 11M                         & 2 epochs                  & \bf 84.0           \\
    \end{tabular}
    \caption{Entity extraction comparison results using DocTr models at different pre-training stages.}
    \label{tab:pretrain}
\end{table}

\newpage

\end{document}